\documentclass[journal]{IEEEtran}

\usepackage{amsmath,amssymb,amsfonts}
\usepackage{graphicx}
\usepackage{booktabs}
\usepackage{multirow}
\usepackage{hyperref}
\usepackage{cleveref}
\usepackage{algorithm}
\usepackage{algpseudocode}
\usepackage{enumitem}
\usepackage[caption=false]{subfig}
\usepackage{xcolor}
\usepackage{colortbl}
\usepackage{tcolorbox}
\tcbuselibrary{skins, breakable}

\begin{document}

\title{Constructing and Evaluating Clinical Reasoning Trajectories for Medical Agent}

\author{%
  \IEEEauthorblockN{Yunqi Zhu\textsuperscript{1,2,*}} \;\;
  \IEEEauthorblockN{Wen Tang\textsuperscript{3}} \;\;
  \IEEEauthorblockN{Jinghao Niu\textsuperscript{1,2}} \;\;
  \IEEEauthorblockN{Wensheng Zhang\textsuperscript{1,2}} \;\;
  \IEEEauthorblockN{Xuebing Yang\textsuperscript{1,2,*}\thanks{*Corresponding authors.\\\indent Emails: zhuyunqi96@163.com; yangxuebing2013@ia.ac.cn}} \\
  \vspace{0.2em}
  \normalsize{ 
  \textsuperscript{1}Guangzhou University, Guangzhou, China\\
  \textsuperscript{2}Institute of Automation, Chinese Academy of Sciences, Beijing, China\\
  \textsuperscript{3}Department of Geriatrics, Beijing Friendship Hospital, Capital Medical University, Beijing, China\\
  }
}

\maketitle

\begin{abstract}
Evaluation of medical artificial intelligence (AI) agents remains predominantly answer-centric,
assessing only the correctness of final outputs while overlooking the quality of intermediate reasoning.
We propose MedTraj, a framework that treats reasoning trajectories as primary objects
for modeling the progressive refinement of clinical evidence toward diagnostic decisions.
A reasoning trajectory captures the state transitions among medical concepts---from clinical observations through evidential reasoning to a final conclusion---making the implicit reasoning chain of a medical agent explicit and inspectable.
Each trajectory is scored across five quality dimensions:
coherence, evidence support, hallucination, completeness, and traceability.
Controlled error injection introduces targeted faults into otherwise correct trajectories
to establish causal links between specific reasoning failures and measurable quality degradation.
Step-level filtering based on marginal contribution
identifies which individual reasoning steps drive or undermine reasoning trajectory quality.
Quality-weighted reasoning trajectory context feeds these evaluations
back into the model at inference time.
Experiments across CareQA, PubMedQA, and CECMed demonstrate that reasoning trajectory context
consistently improves reasoning coherence.
For the severity assessment on the CECMed, quality-weighted reasoning trajectory context nearly doubles the correctness
over the zero-shot baseline while cutting the hallucination ratio by 87\%.
Experiments show that 15.3\% of reasoning steps
carry most of the quality signal,
and that extending chains beyond four steps yields diminishing returns.
Validation by three clinical experts confirms that the automated trajectory quality metric generally aligns with human judgment, demonstrating that fine-grained trajectory decomposition provides an actionable pathway toward more trustworthy medical AI.
\end{abstract}

\begin{IEEEkeywords}
Medical Question Answering, Reasoning Trajectory, Context Learning,
Medical Agent, Large Language Models
\end{IEEEkeywords}

\section{Introduction}

Large language model (LLM)-based Agents have shown remarkable promise in medical question answering,
clinical diagnosis support, and documentation tasks~\cite{singhal2023large, thirunavukarasu2023large, bedi2026holistic, brodeur2026performance}.
As these models are increasingly considered for deployment in clinical workflows,
the focus has shifted from whether they can produce correct answers to whether their reasoning processes 
are trustworthy~\cite{wei2022chain, MedXpertQA2025, mishra2026explainability}.
The medical community has increasingly emphasized that the transparency and explainability of AI-driven reasoning are indispensable for clinical adoption~\cite{reddy2022ExpAIMed},
because trustworthy medical AI must expose not only what it concludes but how it arrives there.
Although multiple reasoning trajectories may converge on the same diagnosis,
a trajectory built on fabricated evidence or incoherent logic remains clinically hazardous:
it may lead to inappropriate treatment plans, misallocation of diagnostic resources, and erosion of clinician trust,
even when the terminal answer happens to be correct.
Detecting such latent reasoning failures, however, requires substantial expert effort~\cite{zhu2025can, llmlackessmetacog2025}.

Currently, the evaluation paradigm in medical artificial intelligence is answer-centric.
Benchmarks such as MedQA~\cite{jin2021medqa}, MedMCQA~\cite{pal2022medmcqa}, and PubMedQA~\cite{jin2019datasetbiomed} measure whether a model selects the right option,
while more recent suites such as MedHELM~\cite{bedi2026holistic} and MedXpertQA~\cite{MedXpertQA2025} expand coverage to broader clinical competencies,
yet the unit of evaluation is still the final answer.
Several reasoning enhancement methods have since emerged to address this gap.
Self-Discover~\cite{zhou2024selfdiscover} constructs reasoning structures by letting models compose primitive reasoning modules,
while retrieval augmented generation methods could ground the model outputs in external medical literature, reducing hallucination at the knowledge level~\cite{raghealthcaresurvey2025, sohn2025rationale}.
Moreover, Self-Consistency~\cite{wang2023self} and Tree-of-Thought~\cite{yao2023tree} strategies improve robustness by aggregating over multiple reasoning paths.
In addition, primarily concentrated in mathematics and code generation,
process reward models and step level verification methods \cite{lightman2023let, zheng2026prmsurvey} provide fine-grained supervision over intermediate steps.
More recently, MediEval~\cite{medieval2025} introduces a multi-dimensional benchmark that evaluates patient contextual and knowledge grounded reasoning,
yet its evaluation granularity still targets aggregated scores rather than individual reasoning steps.

\begin{figure}[t]
  \centering
  \includegraphics[width=1.0\linewidth]{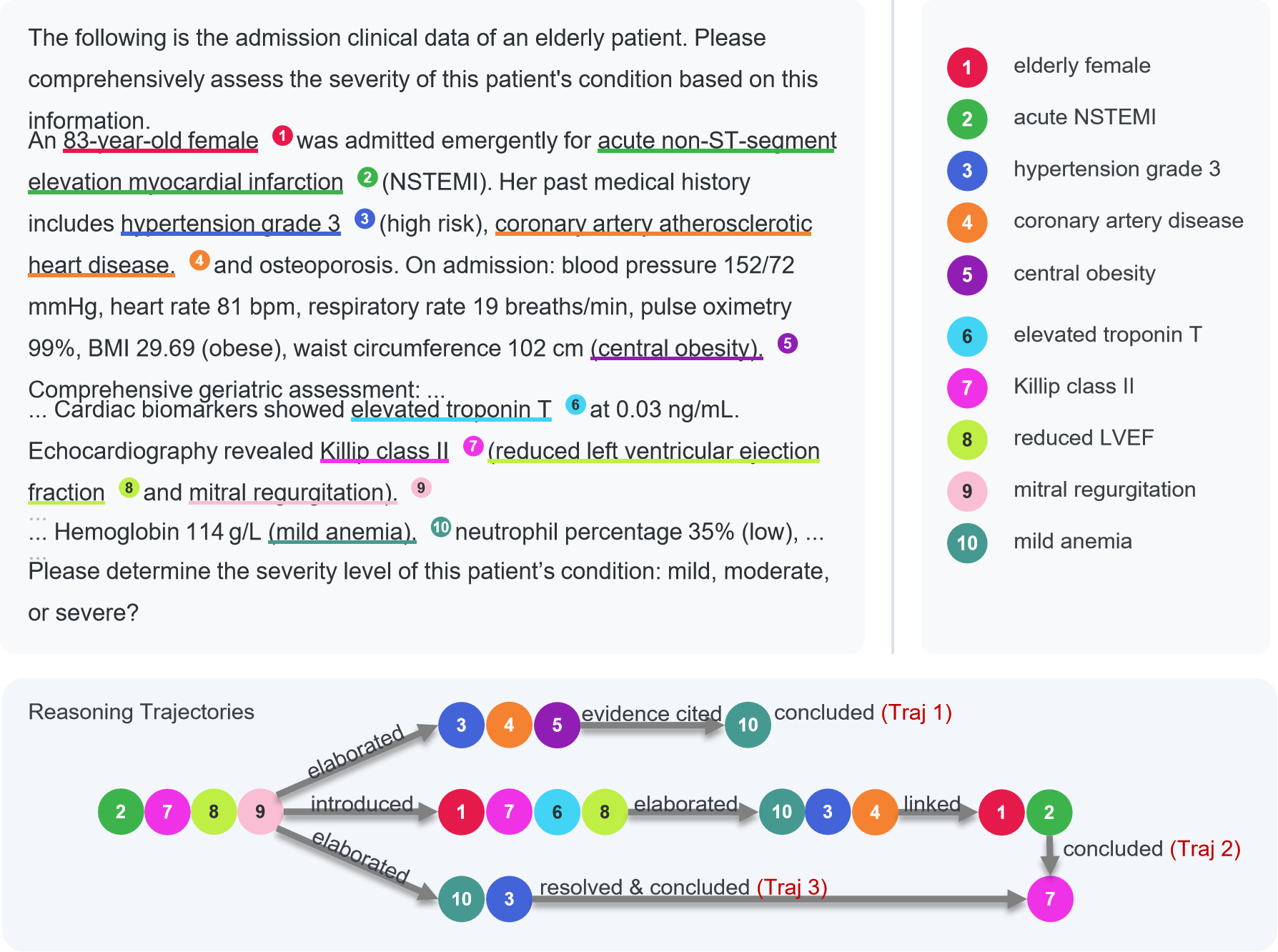}
  \vspace{-0.5em}
  \caption{
    Illustration of medical reasoning trajectories.
  }
  \label{fig:medtraj_trajectory_example}
\end{figure}

\begin{figure*}[h]
  \centering
  \includegraphics[width=0.98\linewidth]{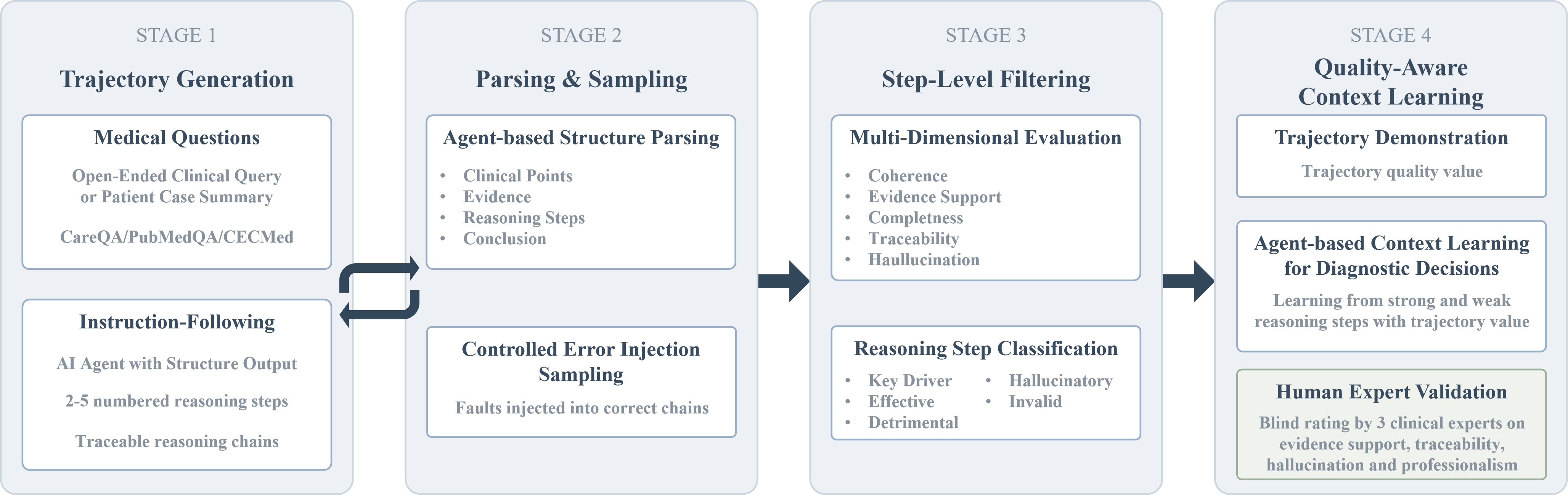}
  \vspace{-0.5em}
  \caption{
    Overall pipeline of MedTraj.
  }
  \label{fig:medtraj_overall_framework}
\end{figure*}

A parallel line of work has explored LLM-based agents and multi-agent frameworks for clinical decision support.
General purpose agent architectures such as ReAct~\cite{yao2023react}, which synergizes reasoning traces with action execution,
have established foundational design patterns adopted by downstream medical systems,
and Voyager~\cite{wang2024voyager}, which learns open ended skills through continuous exploration and self verification.
Agent systems that synergize retrieval, tool use, and reflective self critique can generate extended reasoning trajectories
that resemble the deliberative process of clinical reasoning~\cite{jiang2025medagentbench, chen2025multiagentdiag, zhuohan2026clinicalagents}.
However, evaluating such agents remains challenging:
benchmarks like AgentBench~\cite{liu2024agentbench} show that LLM agents have large performance gaps across environments,
and existing medical benchmarks lack the granularity to assess individual reasoning steps within a trajectory.
Furthermore, an agent may generate a reasoning chain that is logically coherent in surface form yet grounded in fabricated medical claims,
or retrieve context that is topically relevant but clinically misleading.
Therefore, recent studies have begun to move beyond answer correctness alone and prioritize the quality of reasoning trajectories,
with growing interest in constructing and evaluating high quality trajectory datasets~\cite{Fenglitrajsurvey2025}.

As illustrated in Figure~\ref{fig:medtraj_trajectory_example},
given the same clinical context,
different reasoning trajectories generated by the LLM-based medical agent may engage distinct medical concepts
and follow different state transition patterns among them,
ultimately leading to divergent or consistent diagnostic conclusions.
Both the reasoning process and the final outcome carry clinical significance.

Clinicians place equal weight on the diagnostic conclusion and the reasoning process that leads to it~\cite{reddy2022ExpAIMed, chen2025evalhealthcare}.
However, LLM-based medical agents exhibit concerning failure modes: 
hallucinating plausible-looking clinical evidence, 
propagating errors across multi-step reasoning chains, 
and operating without systematic quality assessment of generated trajectories~\cite{omar2025multi, wang2025surveyagents, wu2026medreason}.
These challenges underscore the need for frameworks that can construct reliable reasoning trajectories and diagnose where and why they fail.
Self-reflection has emerged as a promising direction for improving medical reasoning quality.
MedReflect~\cite{huang2026medreflect} enables LLMs to generate single-pass reflection chains for self-verified medical problem-solving,
while Med-REFL~\cite{yang2026medrefl} leverages tree-of-thoughts exploration with automated preference learning to train fine-grained self-correction.
However, the effectiveness of self-reflective reasoning in medical settings remains dataset- and model-dependent~\cite{zhan2026selfcorrect},
indicating a gap between reasoning transparency and reasoning correctness.
Multi-turn evaluation further reveals that LLMs tend to commit to diagnoses prematurely before accumulating sufficient evidence,
and that strategically deferring the diagnostic question to later turns can substantially improve accuracy~\cite{fang2026mint}.

Inspired by these advances, we propose MedTraj, a framework that treats reasoning trajectories as primary objects for modeling the progressive refinement of clinical evidence toward diagnostic decisions.
By capturing the state transitions among medical concepts---from clinical observations through evidential reasoning to a final conclusion---MedTraj makes the implicit reasoning chain of a medical agent explicit and inspectable.
Figure~\ref{fig:medtraj_overall_framework} illustrates the overall pipeline.
First, structured trajectory generation produces multi step clinical reasoning chains from three medical QA sources spanning open ended diagnosis, 
evidence based judgment, and geriatric severity assessment respectively. 
Second, each trajectory is parsed into clinical observations, evidences, numbered reasoning steps, and a final conclusion, 
then scored on five quality dimensions: coherence, evidence support, hallucination, completeness, and traceability. 
Controlled error injection introduces targeted faults into otherwise correct trajectories to establish causal links between specific reasoning failures and measurable quality degradation. 
Next, step level filtering based on marginal contribution identifies which individual steps drive or undermine trajectory quality. 
Further, quality weighted context learning feeds trajectory evaluations back into the model at inference time,
guiding the model to produce more accurate diagnostic decisions by learning from both strong and weak reasoning demonstrations.

Experiments across three medical datasets (CareQA~\cite{arias2025autoeval}, PubMedQA~\cite{jin2019datasetbiomed} and CECMed~\cite{jinwen2026estab}) show that trajectory context consistently improves reasoning quality,
with coherence gains of +0.029 to +0.041 over a zero shot baseline.
Quality weighted context achieves 73.8\% correctness on the severity assessment from CECMed while reducing hallucination by 87\% relative to the baseline.
Step level analysis reveals that only 15.3\% of reasoning steps act as key drivers that disproportionately determine whether a trajectory succeeds,
and that the number of reasoning steps forming a practical upper bound for medical reasoning length may be as few as four.
Validation on CECMed by three Chinese clinical experts confirms that the proposed trajectory value metric exhibits consistent relationships with the ratings assigned by the experts across quality dimensions and statistically significant rank correlations,
establishing that the automated quality assessment captures clinically meaningful distinctions.
The main contributions of this work are:
\begin{itemize}[nosep]
  \item A pipeline for constructing structured medical reasoning trajectories with controlled error injection, 
  enabling causal analysis of how specific fault types degrade reasoning quality.
  \item A multi dimensional trajectory evaluation scheme with a composite trajectory value metric 
  and a step level filtering mechanism that classifies each reasoning step by its marginal contribution.
  \item An experimental study across three medical benchmarks demonstrating that quality weighted trajectory context improves coherence,
  correctness, and hallucination control, providing practical guidelines for constructing and evaluating reasoning trajectories of LLM-based medical agents,
  including evidence-based recommendations on optimal reasoning length and step level quality allocation.
\end{itemize}

\section{Related Work}
\label{sec:related}

\subsection{Medical Reasoning with LLMs}
LLMs have demonstrated substantial capability a broader range of clinical tasks including 
clinical outcome prediction, documentation, and diagnostic dialogue~\cite{thirunavukarasu2023large, zhou2023survey, wang2025llminmed}.
However, studies have shown that LLMs can produce plausible-sounding but factually incorrect clinical narratives,
exhibit overconfidence under uncertainty and missing clinical information~\cite{llmlackessmetacog2025},
and generate hallucinated medical claims that are difficult for non-expert users to detect~\cite{zhu2025can}.

\subsection{Multi-Step Diagnostic Reasoning}
Multi-step diagnostic reasoning is critical to clinical practice and has attracted great attention in recent years.
Chain of thought~\cite{wei2022chain} generates step-by-step clinical reasoning from observations to diagnosis,
improving transparency but remaining vulnerable to error accumulation along a single path.
Tree-of-Thought~\cite{yao2023tree} extends this by exploring multiple hypotheses in parallel and pruning less probable branches.
ReAct~\cite{yao2023react} synergizes reasoning traces with external actions such as knowledge retrieval,
allowing the model to iteratively refine its diagnostic conclusions.
Beyond single-agent approaches, multi-agent collaboration has emerged as a powerful strategy.
MDagents~\cite{kim2024mdagents} assigns specialized medical roles to multiple agents and aggregates their analyses through consensus,
while \cite{ke2024mitigating} further target cognitive biases such as anchoring and premature closure via mutual critique among agents.
KG4Diagnosis~\cite{zuo2025kg4diagnosis} augments hierarchical multi-agent reasoning with medical knowledge graphs to improve factual grounding,
and ClinicalLab~\cite{yan2026clinicallab} aligns agents across multiple departments to simulate multi-disciplinary diagnostic workflows.
Moreover, a complementary direction explores self-reflection as a mechanism for refining the reasoning.
MedReflect~\cite{huang2026medreflect} introduces a single-pass reflection chain comprising hypothesis generation, self-questioning, and decision refinement,
enabling LLMs to improve medical problem-solving without external retrieval.
Med-REFL~\cite{yang2026medrefl} further advances this direction by exploring multiple reasoning paths through tree-of-thoughts search
and constructing preference data via structural assessment to train fine-grained self-correction.

\subsection{In-Context Learning}
In-context learning~\cite{dong2024incontextsurvey} refers to the ability of AI agents to adapt to new tasks 
by conditioning on a small set of input--output examples provided directly in the context.
Its performance is sensitive to the choice and ordering of these demonstrations,
\cite{liu2022good} investigate what makes good in-context examples,
finding that label distribution and input similarity significantly affect performance.
\cite{rubin2022retrieve} propose learning to retrieve task-specific prompts
by training an encoder to select the most relevant demonstrations for each input.
\cite{su2023selective} introduce selective annotation,
choosing a small pool of examples to annotate and retrieving from it at test time to reduce annotation cost.
\cite{wu2023selfadaptive} further adapt example selection and ordering
from an information compression perspective, allowing each input to find its optimal demonstration permutation.
Recently, \cite{agarwal2024manyshot} demonstrate that scaling demonstrations
to hundreds or thousands within the context window yields consistent gains on complex reasoning tasks.
In the medical domain, \cite{sivarajkumar2024prompting} systematically evaluate zero-shot and few-shot prompting strategies across five clinical tasks,
\cite{wu2024knowlab} investigate few-shot chain-of-thought prompting for medical error detection and correction,
showing that the quality and ordering of clinical demonstrations directly affect the model's ability to identify reasoning errors in clinical notes.

\subsection{Evaluation of Medical LLMs}
Conventional medical benchmarks~\cite{jin2021medqa, pal2022medmcqa, jin2019datasetbiomed} 
assess the AI agent through the reference question and answer,
but do not capture the sequential and interactive nature of clinical decision-making.
Workflow-based simulation benchmarks attempt to address this gap:
ChatCoach~\cite{huang2024chatchoach} uses LLMs themselves to evaluate the agent communication and decision-making quality in patient consultations;
MedChain~\cite{liu2025medchain} contains over 12k cases across 19 specialties with multi-step dialogue-based diagnosis;
AgentClinic~\cite{schmidgall2026agentclinic} simulate a clinical environment for LLM agents, offering both multimodal and dialogue-based clinical evaluation;
and ClinicalLab~\cite{yan2026clinicallab} tests diagnostic performance across 24 departments and 150 diseases.
LLM-as-judge has emerged as a scalable alternative,
where a stronger agent scores the coherence, evidence support, and hallucination of generated trajectories,
though its reliability depends on the judge's own clinical competence.
CliniCARE-Bench~\cite{chatrath2026clinicare} introduces a clinical calibrated audit that evaluates medical reasoning directly within EHR workflows,
and MTDiag~\cite{chouayfati2026mtdiag} proposes a multi-turn diagnostic dataset designed for clinically meaningful evaluation of LLM reasoning beyond single-turn accuracy.
MINT~\cite{fang2026mint} introduces a multi-turn benchmark with clinically labeled evidence shards for studying incremental evidence accumulation,
the AI agents tend to commit to diagnoses prematurely and that deferring the diagnostic question can substantially improve accuracy.

\section{Method}
\label{sec:method}

\subsection{Overview}

In this work, we propose MedTraj,
a framework that treats reasoning trajectories as primary objects for modeling the progressive refinement of clinical evidence toward diagnostic decisions.
By capturing the state transitions among medical concepts,
MedTraj makes the implicit reasoning chain of a medical agent explicit and inspectable,
enabling systematic quality assessment beyond answer correctness.
Controlled error injection establishes causal links between specific reasoning failures and measurable quality degradation,
while step-level filtering identifies which individual steps drive or undermine trajectory quality.
The pipeline proceeds in the following stages:
(i)~trajectory generation from medical QA benchmarks via a large instruction-following model with
structured output;
(ii)~structured parsing, controlled error injection, and multi-dimensional
quality evaluation;
(iii)~step-level filtering based on marginal contribution analysis;
and (iv)~quality-aware context learning that guides the model toward more accurate diagnostic decisions
by feeding trajectory evaluations back at inference time.


\subsection{Data Preparation}
\label{sec:data}

In this work, we focus on three medical datasets. 
CareQA~\cite{arias2025autoeval} contains open-ended clinical questions sourced from MedQA, MedMCQA, HeadQA, and MMLU. 
We use it to represent free-form diagnostic reasoning. 
PubMedQA~\cite{jin2019datasetbiomed} contains open-ended clinical questions derived from PubMed article abstracts, 
requiring the model to produce a reasoned answer supported by biomedical evidence.
CECMed~\cite{jinwen2026estab} targets severity assessment for geriatric patients,
a specialized clinical task in which the model takes a patient's admission and medical history report summary as input
and predicts the severity level as judged by clinical experts.

For each dataset we randomly sample 2{,}000 examples for trajectory generation and
model training, and hold out 500 examples for evaluation (CECMed: 1{,}908 training and 500 test). 
The sample sizes are chosen to balance computational
cost with statistical reliability. Every record is stored in a uniform format
comprising an item index, dataset identifier, task type, question, and reference answer. 


\subsection{Reasoning Trajectory Generation}
\label{sec:generation}

Given a medical question, we ask an agent to produce a structured clinical reasoning trajectory.  
The prompt instructs the model to reason as a clinician:
enumerate relevant clinical observations, cite supporting medical evidence, 
lay out a step-by-step reasoning chain, and state a final conclusion.  
We require the output to contain two to five numbered
reasoning steps so that the trajectory is both traceable and amenable to step-level analysis.

Rather than generating a single response per question, we sample multiple
trajectories at varied temperatures and repeat each configuration $n$ times.  
This produces several reasoning paths for the same question,
providing raw material for the trajectory recombination and comparative
analysis described below.  Each generated trajectory is paired with its
source question and the reference answer, forming a
(question, trajectory, answer) triple that serves as the basic unit for
all subsequent evaluation and training.


\subsection{Structured Parsing and Quality Dimensions}
\label{sec:parsing}

Raw trajectory text must be decomposed into analyzable components before
we can evaluate or recombine it.  We parse each trajectory into four
standardized elements: clinical points (key observations extracted
from the question), evidence (cited medical knowledge),
reasoning steps (numbered intermediate inferences), and
conclusion (the final diagnosis or answer).

This decomposition reveals six structural types that trajectories can take,
ranging from minimal to fully elaborated, as summarized in Table~\ref{tab:struct_types}.

\begin{table}[htbp]
  \centering
  \caption{Structural types of parsed trajectories.}
  \label{tab:struct_types}
  \small
  \scalebox{0.65}{
  \begin{tabular}{lp{8.2cm}}
    \toprule
    Type & Description \\
    \midrule
    Points Only & Clinical observations with no reasoning or conclusion \\
    Points+Reasoning & Observations and reasoning but no conclusion \\
    Reasoning+Result & A reasoning chain without explicit point extraction \\
    Points+Reasoning+Result & The complete observation-to-conclusion chain \\
    Points+Evidence+Reasoning+Result & Observation-to-conclusion chain with explicit evidence citations \\
    Points+Reasoning+Subreasoning & Hierarchical chain with nested sub-inferences \\
    \bottomrule
  \end{tabular}
  }
\end{table}

Parsing is carried out by an LLM parser that handles ambiguous or free-form outputs
where keyword matching is insufficient. 
With parsed trajectories in hand, we define five quality dimensions:

\begin{enumerate}[nosep]
  \item Coherence: how well each reasoning
        step connects to the clinical points and to adjacent steps.
  \item Evidence support: the degree to which
        claims are grounded in cited medical evidence.
  \item Hallucination: the proportion of assertions that are fabricated or unsupported by
        established medical knowledge.
  \item Completeness: whether the reasoning chain
        covers all necessary steps from observation to conclusion.
  \item Traceability: whether the final conclusion
        can be logically derived from the preceding reasoning.
\end{enumerate}

The weights reflect the judgment about which failure modes matter most
in clinical settings: hallucinated medical claims carry the heaviest
penalty because they can lead to harmful decisions, 
while coherence receives the highest positive weight because it captures the
backbone of sound clinical reasoning.


\subsection{Controlled Error Injection}
\label{sec:injection}

To understand how specific error types degrade reasoning quality, and to
build negative samples, we implemented seven controlled injection
modes that introduce targeted faults into otherwise correct trajectories
(Table~\ref{tab:injection}).

\begin{table*}[htbp]
  \centering
  \caption{Error injection modes.}
  \label{tab:injection}
  \small
  \scalebox{1.0}{
  \begin{tabular}{@{}p{3.4cm}p{10.0cm}p{3.0cm}@{}}
    \toprule
    Mode & Description & Error type \\
    \midrule
    Inject irrelevant & Insert steps from another question's trajectory at a random position & Redundant reasoning \\
    Swap steps & Shuffle the order of selected reasoning steps & Structural disruption \\
    Flip numeric & Multiply numerical values by a random factor uniformly drawn from 2 to 10 & Factual error \\
    Remove key point & Delete critical steps (retaining $\geq 50\%$) & Incompleteness \\
    Replace unrelated & Replace steps with another question's reasoning & Causal contradiction \\
    Wrong-then-correct & Prepend $N$ erroneous steps before the correct chain & Positional: bad prefix \\
    Correct-then-wrong & Append $N$ erroneous steps after the correct chain & Positional: bad suffix \\
    \bottomrule
  \end{tabular}
  }
\end{table*}

First, we distinguish structural disruptions (inject irrelevant, swap steps, and
replace unrelated) that alter the reasoning chain's organization from
local perturbations (flip numeric and remove key point) that
modify individual steps.  
Second, the wrong-then-correct and
correct-then-wrong modes test the effect of error position:
does a misleading prefix cause more damage than a misleading suffix, or vice versa? 
After injection, each corrupted trajectory is re-evaluated on all quality dimensions.  
The resulting (error type, quality drop) pairs
establish causal, rather than merely correlational, links between
specific reasoning faults and measurable quality degradation.


\subsection{Multi-Dimensional Evaluation}
\label{sec:evaluation}

Each trajectory is scored by an LLM evaluator that
assesses all five quality dimensions plus two additional metrics:
average step score, which measures the proportion of
clinically relevant tokens within each step and the semantic coherence between adjacent steps; 
answer match score, which quantifies agreement between the trajectory's conclusion and the reference answer.

These per-dimension scores are aggregated into a single trajectory
value through a weighted combination of positive and penalty terms.
The positive component sums six quality dimensions:
\begin{equation}
  \label{eq:splus_value}
  S^{+} = \sum_{k \in \mathcal{K}^{+}} w_{k} \, s_{k},
\end{equation}
where $\mathcal{K}^{+}$ indexes the six positive dimensions: coherence, evidence, step, completeness, traceability, and match.
The penalty component combines two terms:
\begin{equation}
  \label{eq:sminus_value}
  S^{-} = \sum_{j \in \mathcal{K}^{-}} w_{j} \, s_{j},
\end{equation}
where $\mathcal{K}^{-} $ indexes negatiove dimensions: hallucination, invalid steps.
Both components are normalized so that their internal weights sum to unity.  
The final trajectory value is obtained by passing the difference
$S = S^{+} - S^{-}$ through a sigmoid, mapping $V$ to $[0, 1]$.

\begin{equation}
  \label{eq:traj_value}
  V = \frac{1}{1 + e^{- S \cdot \kappa}},
\end{equation}


\subsection{Step-Level Filtering and Classification}
\label{sec:filtering}

Trajectory-level scores treat every reasoning step as contributing equally, 
but in practice some steps matter far more than others.  
To capture this heterogeneity, we perform step-level analysis based on
marginal contribution.

We identify trajectory pairs that share a common prefix, 
for example, trajectory~$i$ follows steps $[a, b, c]$ while trajectory~$j$ extends
the same prefix with $[a, b, c, k]$.  
By comparing the trajectory
values of such pairs, we isolate the marginal effect $\delta$ of the
divergent step.

Each step is then classified into one of five categories based on its
marginal~$\delta$ and the answer match score (MS) of its host
trajectory:

\textbf{Key driver} ($\delta \geq +0.05$, MS$\geq 0.8$):
      the step substantially lifts the trajectory toward a correct,
      high-quality outcome.

\textbf{Effective} ($\delta < +0.05$, MS$\geq 0.8$):
      the step maintains an already-high quality level without
      dramatic improvement.

\textbf{Detrimental} ($\delta < -0.02$):
      the step actively degrades cumulative quality, regardless of
      the final answer.

\textbf{Hallucinatory} ($\delta \geq +0.05$, MS$< 0.8$):
      the step appears locally informative but the overall trajectory
      remains inaccurate, a plausible-sounding but misleading
      inference.

\textbf{Invalid} ($\delta < +0.05$, MS$< 0.8$):
      a filler step in a low-quality trajectory that contributes
      neither positively nor negatively.

Three findings emerge from this classification.  
First, key driver steps account for only 15.3\% of all steps yet disproportionately
determine whether a trajectory succeeds or fails.  
Second, detrimental steps (18.5\%) are nearly as common, 
indicating that most reasoning chains contain at least one point where quality breaks down.  
Third,
position matters: step~2 exhibits the highest mean~$\delta$
(+0.261) and the largest variance, with key drivers and hallucinatory
steps together comprising 69\% of all step-2 labels.  
This suggests that the second reasoning step is a critical decision point: 
it either sets the reasoning direction firmly on track or introduces a deviation
that propagates through the rest of the chain.  
Steps~3--5 show stable,
gradually diminishing contributions, while step~6 and beyond exhibit a
sharp quality collapse (Section~\ref{sec:length}).


\subsection{Modeling Concept State Transitions in Reasoning}
\label{sec:concept_transition}

Beyond trajectory-level quality scores, we aim to characterize how the reasoning unfolds by modeling the progression of medical concepts across a trajectory.
Given an input question~$q$ and a reasoning trajectory~$\tau = (s_1, s_2, \ldots, s_n)$ composed of $n$ sentences, we extract and track the evolution of core medical concepts through a three-stage pipeline.

\textbf{Concept extraction: }
We first employ an LLM to identify a set of core medical concepts from the question,
$\mathcal{C}_q = \{c_1, c_2, \ldots, c_K\}$,
where each $c_k$ denotes a clinically salient entity such as a symptom, laboratory finding, or demographic attribute.
For each sentence $s_i$ in the trajectory, the same model determines which concepts from $\mathcal{C}_q$ it engages and whether it introduces novel concepts $\mathcal{C}_{\text{novel}}^{(i)}$ not present in the original question.
The full concept set at sentence $s_i$ is therefore $\mathcal{C}^{(i)} \subseteq \mathcal{C}_q \cup \mathcal{C}_{\text{novel}}^{(i)}$.

\textbf{State assignment: }
For each sentence $s_i$ and its associated concept $c \in \mathcal{C}^{(i)}$, we assign a reasoning state $z_i \in \mathcal{Z}$ from a predefined taxonomy of 8 cognitive states:
\textit{introduced} (first mention of $c$),
\textit{evidence cited} (clinical data or literature invoked as evidence for $c$),
\textit{elaborated} (further explanation or reasoning about $c$),
\textit{linked} (associating $c$ with another concept via causal, conjunctive, or contrastive relations),
\textit{contradicted} (challenging or negating $c$),
\textit{resolved} (drawing a conclusive judgment about $c$),
\textit{concluded} (deriving a final diagnostic decision grounded in $c$),
and \textit{irrelevant} (the sentence does not engage any core concept).
Formally, the state assignment for sentence $s_i$ is:
\begin{equation}
  z_i = f_{\text{state}}(s_i,\, \mathcal{C}^{(i)};\, \theta_{\text{LLM}}),
\end{equation}
where $f_{\text{state}}$ is parameterized by the LLM.

\textbf{Transition scoring: }
The discrete state labels $z_i$ alone do not capture the degree of semantic change between consecutive reasoning steps.
We therefore complement them with a lightweight sentence embedding model $g_\phi$ to obtain a continuous measure of semantic proximity:
\begin{equation}
  \sigma_{i,i+1} = \cos(g_\phi(s_i), g_\phi(s_{i+1})).
\end{equation}
Each state transition $(z_i, z_{i+1})$ is then validated against $\sigma_{i,i+1}$:
a transition is accepted as the state pair follow the reasonable progression and the embedding similarity exceeds a threshold~$\gamma$, filtering out cases where the LLM-assigned states jump between unrelated content.
We define a hierarchy over the 8 states,
and classify each consecutive pair as ascending, stable, or descending.
The concept trajectory transition score aggregates over all $n{-}1$ consecutive transitions:
\begin{equation}
  S_{\text{trans}}(\tau) = \frac{1}{n-1} \sum_{i=1}^{n-1} \sigma_{i,i+1} \cdot \mathbb{I}\bigl[\sigma_{i,i+1} > \gamma \;\wedge\; (z_i, z_{i+1}) \in \mathcal{P}\bigr],
\end{equation}
where $\mathcal{P}$ is the set of state pairs.


\subsection{Quality-Aware Training}
\label{sec:training}

\definecolor{myc1}{RGB}{173, 197, 222}
\definecolor{myc2}{RGB}{200, 200, 200}
\begin{table*}[htbp]
  \centering
  \caption{
    LLM evaluation results across three datasets.
    Blue marks the best method and grey marks the second-best method.
  }
  \label{tab:main_results}
  \small
  \scalebox{1.0}{
  \begin{tabular}{@{}lcccc|cccc|cccc@{}}
    \toprule
    & \multicolumn{4}{c}{CareQA (Open-Ended)}
    & \multicolumn{4}{c}{PubMedQA (Open-Ended)}
    & \multicolumn{4}{c}{CECMed (Severity)} \\
    \cmidrule(lr){2-5} \cmidrule(lr){6-9} \cmidrule(lr){10-13}
    Method      & Corr. & Coh. & Evi. & Hal. & Corr. & Coh. & Evi. & Hal. & Corr. & Coh. & Evi. & Hal. \\
    \midrule
    NC          & 0.229 & 0.538 & 0.430 & 0.523 & 0.656 & 0.812 & 0.800 & 0.140 & 0.390 & 0.734 & 0.692 & 0.189 \\
    QA          & 0.232 & 0.540 & 0.429 & 0.546 & 0.651 & 0.809 & 0.798 & 0.147 & 0.436 & 0.745 & 0.705 & 0.176 \\
    SFT         & 0.220 & 0.549 & 0.437 & 0.515 & 0.656 & 0.820 & 0.814 & 0.145 & 0.408 & 0.748 & 0.704 & 0.175 \\
    SC          & 0.206 & 0.539 & 0.423 & 0.539 & 0.679 & 0.838 & \cellcolor{myc2}0.837 & \cellcolor{myc2}0.112 & 0.414 & 0.731 & 0.678 & 0.203 \\
    BoN         & 0.238 & 0.565 & \cellcolor{myc1}0.461 & 0.539 & 0.688 & \cellcolor{myc2}0.851 & 0.830 & 0.114 & 0.396 & 0.729 & 0.674 & 0.208 \\
    TC          & \cellcolor{myc2}0.244 & \cellcolor{myc2}0.567 & 0.450 & \cellcolor{myc2}0.510 & \cellcolor{myc2}0.708 & \cellcolor{myc1}0.853 & \cellcolor{myc1}0.851 & \cellcolor{myc1}0.104 & \cellcolor{myc2}0.642 & \cellcolor{myc2}0.773 & \cellcolor{myc2}0.767 & \cellcolor{myc2}0.063 \\
    QWC         & \cellcolor{myc1}0.246 & \cellcolor{myc1}0.576 & \cellcolor{myc2}0.459 & \cellcolor{myc1}0.507 & \cellcolor{myc1}0.730 & \cellcolor{myc1}0.853 & \cellcolor{myc1}0.851 & \cellcolor{myc1}0.104 & \cellcolor{myc1}0.738 & \cellcolor{myc1}0.799 & \cellcolor{myc1}0.797 & \cellcolor{myc1}0.025 \\
    \bottomrule
  \end{tabular}
  }
\end{table*}

We compare the following strategies for feeding trajectory quality information
back into the base LLM model.

For context learning,
at inference time, we prepend $k$ randomly selected examples to the target
question and ask the base LLM model to generate its own reasoning trajectory.  
We compare four context
selection strategies:

No Context (NC): the model receives only the question,
with no demonstration examples.

QA Context (QA): $k$ question--answer pairs are sampled
from the training set, providing answer demonstrations but no
reasoning process.

Trajectory Context (TC): $k$ complete reasoning
trajectories with high trajectory values are sampled from the
evaluation pool, showing the model both how to reason and what
a good answer looks like.

Quality-Weighted Context (QWC): trajectories are
sampled with balanced representation of high- and low-quality
examples, each accompanied by its trajectory value score.
Presenting both extremes lets the model learn what
distinguishes good reasoning from bad.

For supervised fine-tuning (SFT),
we fine-tune the base model on randomly sampled trajectory data. 
Unlike the context learning variants, 
SFT modifies the model weights rather than conditioning on in-context examples.  
We include SFT to assess whether simply exposing
the model to trajectory-format data during training yields benefits
comparable to providing structured demonstrations at inference time.

For inference-time baselines,
we additionally test two sampling-based strategies that do not require
any training data curation: Self-Consistency (SC), which draws
$k$ samples and takes a majority vote on the final answer, 
and Best-of-N (BoN), which draws $k$ samples and selects the
trajectory with the highest trajectory value $V$ in Eq.~\eqref{eq:traj_value}.  
These baselines help disentangle the benefit of what the model sees in context 
from the benefit of how many responses it generates.

\section{Experiments}
\label{sec:experiments}


\subsection{Experimental Setup and Evaluation}
\label{sec:setup}

All experiments use DeepSeek-R1-Distill-Qwen-7B~\cite{modelds1disqwen7b} (7.6\,B parameters, 32 layers) as the
base LLM model for inference and fine-tuning.  Reasoning trajectories are
generated by Qwen3-32B~\cite{modelqwen332b} (32\,B parameters, 64 layers) and evaluated by
Qwen3-8B~\cite{modelqwen38b} (8\,B parameters, 32 layers) acting as the LLM judge.
For all generation, we set the maximum number of new tokens to 4096, the temperature to 0.3, and the top-$p$ sampling threshold to 0.9. 
The sigmoid steepness $\kappa$ is set to 10.
The trajectory value weights $w_k$ in Eq.~\eqref{eq:splus_value} are set to
0.30 for coherence, 0.20 for evidence, 0.20 for step, 0.10 for completeness, 0.10 for trace, and 0.10 for match;
the penalty weights in Eq.~\eqref{eq:sminus_value} are set to
0.60 for hallucination and 0.40 for invalid steps.
We compare seven methods across three datasets.
For context-based methods and sampling-based methods, we uniformally set $k = 5$.

\begin{enumerate}[nosep]
  \item No Context (NC): zero-shot baseline.
  \item QA Context (QA): $k$ question--answer pairs as context.
  \item Trajectory Context (TC): $k$ high-quality trajectories.
  \item Quality-Weighted Context (QWC): balanced high/low-quality
        trajectories with scores.
  \item SFT: supervised fine-tuning on random trajectories.
  \item Self-Consistency (SC): $k$ samples with majority vote.
  \item Best-of-N (BoN): $k$ samples, reward model selects best.
\end{enumerate}

For agent-based evaluation, we report four metrics: correctness, coherence,
evidence score, and hallucination ratio.
All metrics are computed by the LLM evaluator unless otherwise noted.
Hallucination is the lower the better, while the other three metrics are higher-the-better.

Furthermore, to validate the proposed trajectory quality assessment, we conducted an independent expert evaluation study.
Three geriatric medicine experts, each with over 5 years of clinical experience, were recruited to evaluate the reasoning trajectories.
For each of the 100 randomly sampled clinical questions, the experts evaluated 3 randomly sampled model-generated answers that correspond to different levels of trajectory value: high, medium, and low.
Therefore, each question contributed 3 trajectories with distinct quality levels, yielding a total of 300 trajectory evaluations per expert.
The experts were blinded to the trajectory value scores and were asked to rate each trajectory on four dimensions: evidence support, reasoning traceability, hallucination severity, and overall professionalism, using a 5-point Likert scale.


\subsection{Main Results}
\label{sec:main_results}

Table~\ref{tab:main_results} presents the experimental results. 
Additionally, a boxplot visualization of the per-sample metric distributions is provided in
\ref{app:visual} Figure~\ref{fig:appendix_boxplots}.

\textbf{Trajectory context improves reasoning quality across
all the medical QA datasets.}  TC raises coherence by +0.029 on CareQA,
+0.041 on PubMedQA, and +0.039 on CECMed (all vs.~NC).  
Evidence scores improve by +0.020, +0.051, and +0.075 respectively.  
The gain holds whether the task is open-ended QA or severity assessment, 
indicating that observing a well-formed reasoning chain generalizes beyond any
single answer format.

\textbf{Correctness gains depend on task structure.}  
For CECMed, TC lifts correctness from 0.390 to 0.642 (+25.2\%), 
the largest absolute gain in the table. 
For PubMedQA, the gain is smaller but consistent (+0.052).  
For CareQA, correctness barely moves (+0.015) despite clear coherence and evidence improvements.  
This suggests that in open-ended QA, better reasoning does not automatically translate into
a more precise answer, and the model may reason well about the wrong diagnosis.

\subsection{Step-Level Analysis}
\label{sec:step_analysis}

\textbf{QA-only context is generally ineffective.}  QA matches or
slightly trails NC on every metric across all three datasets.  
For CareQA, QA even increases the hallucination (0.546 vs.~0.523).
Switching from QA to TC yields correctness gains of +1.2\%, +5.8\%, and
+20.6\% on the three datasets.  Showing the model how to reason
matters far more than showing it what the answer is.

\textbf{SFT trades correctness for hallucination control.}
For PubMedQA, SFT matches NC on correctness (0.656) but does not reduce hallucination (0.145 vs.~0.140).  
For CECMed, SFT achieves 0.408 correctness, 
only marginally above NC, while its hallucination (0.175) remains close to the baseline.  
Supervised fine-tuning on random trajectories does not selectively reinforce high-quality reasoning.

\begin{figure*}[htbp]
  \centering
  \includegraphics[width=0.98\linewidth]{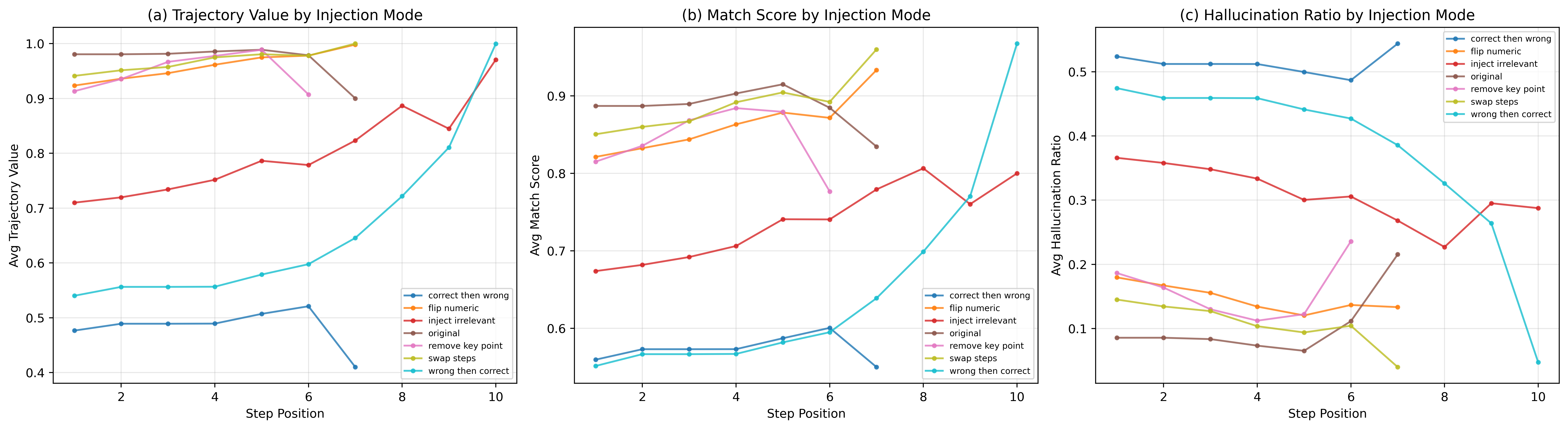}
  \vspace{-0.5em}
  \caption{Growth curves of trajectory value, answer score, and
    hallucination under different error types.
  }
  \label{fig:error_injection_curves}
\end{figure*}

\textbf{Sampling-based methods offer limited gains.}  SC
and BoN produce modest correctness improvements on PubMedQA (+0.023 and +0.032 over NC) 
but fail to move the needle on CareQA and CECMed.  
SC actually lowers correctness on CareQA (0.206 vs.~0.229).  
Simply generating more responses and voting or ranking does not substitute for
providing the model with a structured reasoning demonstration.

\section{Discussion}
\label{sec:discussion}

\subsection{Effect of Trajectory Quality Weighting}
\label{sec:quality_weighting}

QWC extends TC by mixing high- and low-quality trajectories and attaching a quality score to each.  
The effect of this weighting is task-dependent.

For CECMed, QWC delivers the strongest result in the entire table:
correctness reaches 0.738 (vs.~0.642 for TC, +9.6 percentage points),
and hallucination drops to 0.025 (vs.~0.063 for TC, a further 60\% reduction).  
In a specialized clinical domain where the cost of a
hallucinated assertion is high, quality-weighted context provides
substantial additional value over plain trajectory context.

\begin{figure*}[htbp]
  \centering
  \includegraphics[width=0.98\linewidth]{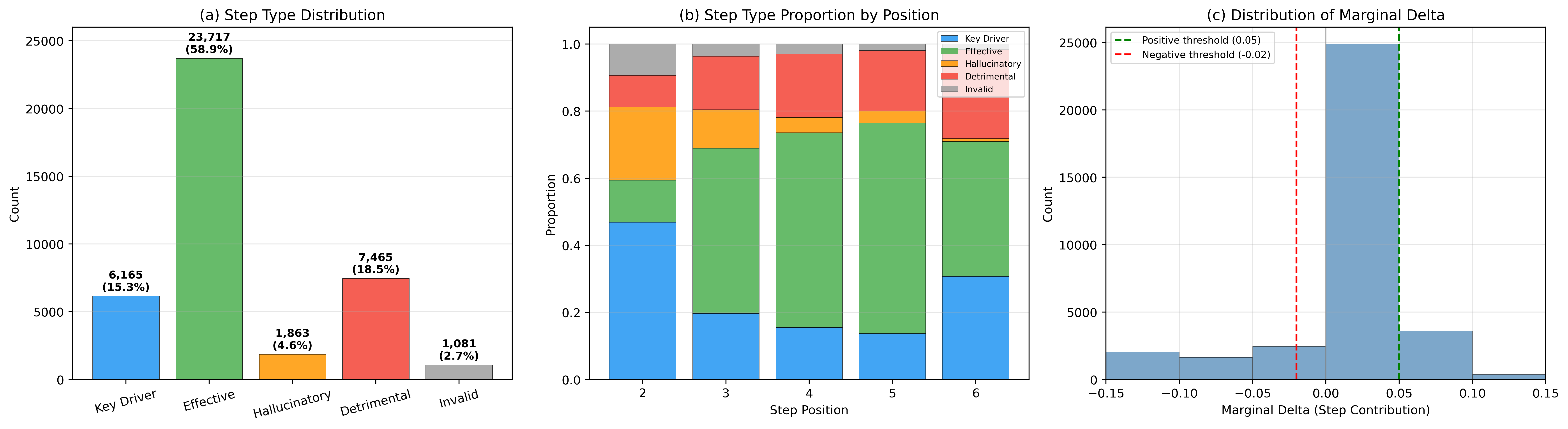}
  \vspace{-0.5em}
  \caption{
    (a)~Step type distribution.  (b)~Proportion of each step type by
    position.  (c)~Distribution of marginal contribution~$\delta$.
  }
  \label{fig:step_distribution}
\end{figure*}

For CareQA, QWC achieves the highest coherence (0.576) and evidence
score (0.459) among all methods, yet its correctness (0.246) is only
0.2 percentage points above TC (0.244).  
This decoupling between reasoning quality and answer correctness is consistent with the
non-linear relationship we observed in Section~\ref{sec:main_results}:
in open-ended QA, a more coherent reasoning chain does not guarantee a
more accurate final answer.

For PubMedQA, QWC and TC are nearly identical across all four metrics,
suggesting that once the context contains good trajectories, 
adding quality labels and low-quality counterexamples provides little extra
signal for open-ended QA tasks.


\subsection{Inference-Time Strategies}
\label{sec:inference}

Self-Consistency (SC) and Best-of-N (BoN) generate multiple responses per question.
SC takes a majority vote on the final answer, 
while BoN selects the trajectory with the highest value score. 
Both methods leave the model unchanged, 
they only adjust the number of generated responses and the selection rule.

On PubMedQA, SC reaches 0.679 correctness and BoN reaches 0.688, 
both above NC (0.656) but below TC (0.708) and well below QWC (0.730).  
On CareQA, SC drops to 0.206, below every other method including NC.  
On CECMed, SC (0.414) and BoN (0.396) both trail NC (0.390) or match it at best, 
while their hallucination ratios (0.203 and 0.208) are the highest in the table.

Sampling more responses helps on PubMedQA
(+0.023 and +0.032 over NC for SC and BoN), where the answer space is
relatively constrained and majority voting can filter out outlier errors.  
But on CareQA and CECMed, where the answer space is
open-ended or where hallucination is the dominant failure mode, 
generating more responses simply produces more hallucinations.
Trajectory context, by contrast, shapes the reasoning process itself
and is effective regardless of answer-space size.


\subsection{Error Injection Analysis}
\label{sec:error_analysis}

In this section, we analyze the controlled error injection experiments to identify which reasoning faults most
severely degrade trajectory quality and to quantify the magnitude of each degradation. 
Figure~\ref{fig:error_injection_curves} shows the growth curves of trajectory value, answer score, 
and hallucination under different error types.
Boxplot distributions of all seven evaluation dimensions under each injection mode are provided in
\ref{app:visual} Figure~\ref{fig:injection_mode_boxplots}.

Structural disruption dominates local perturbation.
Injecting irrelevant steps and replacing steps with another question's
reasoning reduce trajectory value by 27--51\%.  
By contrast, flipping numerical values and removing key points cause only 4--7\% drops.  
The reasoning chain is more fragile to reorganization than to individual factual errors.

The correct-then-wrong
mode produces a trajectory value of 0.477, while wrong-then-correct
yields 0.540.  Trailing errors hurt more than leading errors.  
This suggests that the tail of a reasoning chain exerts disproportionate
influence on the overall quality judgment, possibly because evaluators
(both human and LLM) weight recent information more heavily, 
or because a corrupted conclusion undermines the entire chain retroactively.


\begin{figure*}[htbp]
  \centering
  \includegraphics[width=0.98\linewidth]{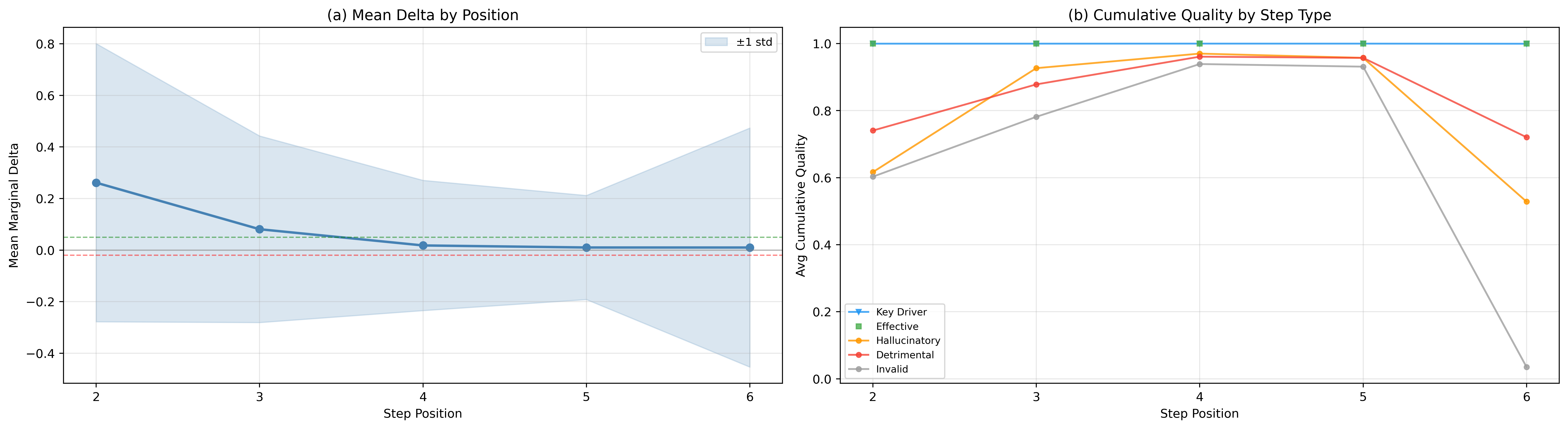}
  \vspace{-0.5em}
  \caption{
    (a)~Mean marginal contribution~$\delta$ by step position
    ($\pm$1 std).  (b)~Cumulative quality curves by step type.
    }
  \label{fig:delta_position}
\end{figure*}

In this section, 
we examine how individual reasoning steps contribute to trajectory quality.  
We apply the prefix-matching procedure described in Section~\ref{sec:filtering} 
to isolate the marginal effect of each step.

Figure~\ref{fig:step_distribution} summarizes the overview of step-type filtering and classification.
Effective steps dominate (58.9\%), maintaining an already-high quality
level without dramatic change.  
Key driver steps (15.3\%) and detrimental steps (18.5\%) together comprise one-third of all steps,
confirming that reasoning chains are not uniformly valuable: 
a minority of steps carry most of the quality signal.

Figure~\ref{fig:delta_position}(a) shows that step~2 has the highest
mean~$\delta$ (+0.261) and the widest spread (std~0.54).  
At this position, 47\% of steps are key drivers and 22\% are hallucinatory,
together accounting for 69\% of all step-2 labels.  
The second reasoning step is where the trajectory either locks onto a correct direction or veers off course.

Steps~3--5 form the stable body of the chain: effective steps comprise 49--63\%, 
detrimental steps hold steady at 16--19\%, 
and mean~$\delta$ decays gradually from +0.081 to +0.010.  
At step~6, an interesting rebound occurs: 
key driver proportion jumps from 13.7\% (step~5) to 30.7\%, 
but detrimental steps also rise from 18.0\% to 26.6\%.  
Step~6 is a high-variance fork: 
it can either rescue a weakening chain or accelerate its collapse.

Figure~\ref{fig:delta_position}(b) shows cumulative quality curves by step type.  
Key driver steps produce the steepest ascent, detrimental steps the steepest descent, 
and effective steps a gentle, sustained climb.  
The divergence between these curves widens with each additional step, 
reinforcing the conclusion that step-level composition matters more than step count alone.


\subsection{Reasoning Length Analysis}
\label{sec:length}

\begin{figure*}[htbp]
  \centering
  \includegraphics[width=0.98\linewidth]{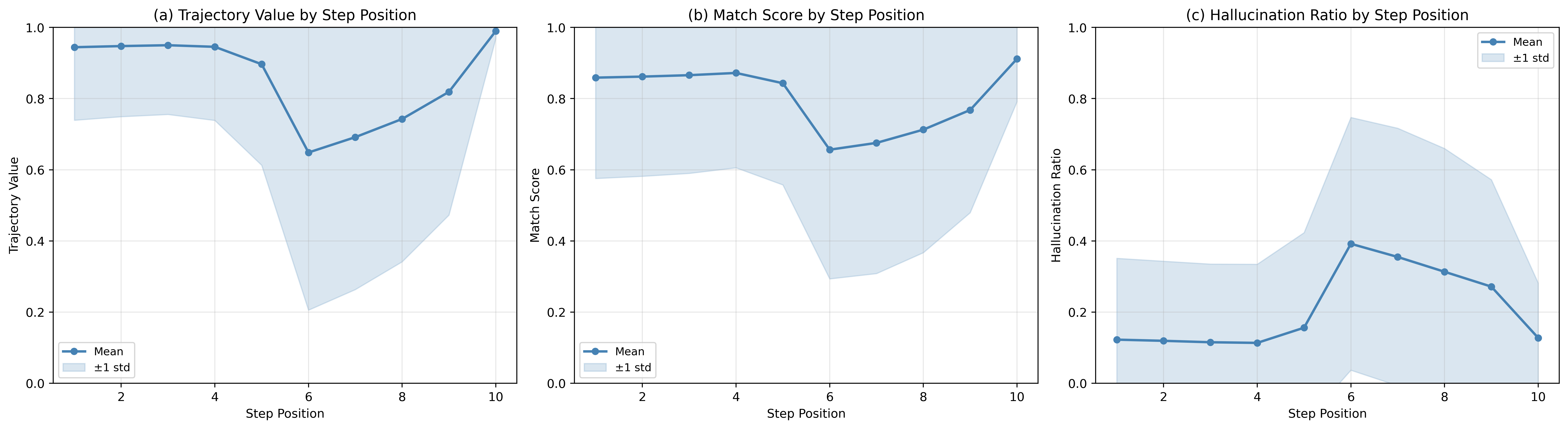}
  \vspace{-0.5em}
  \caption{
    Trajectory value, answer score, and hallucination ratio as
    reasoning steps accumulate (mean $\pm$ std band).
    }
  \label{fig:progressive_quality}
\end{figure*}

In this section, we investigate the reasonable length of a medical reasoning trajectory.
Figure~\ref{fig:progressive_quality} tracks three metrics as steps
accumulate.

Through steps~1--4, trajectory value holds steady at 0.94--0.95, answer
score rises gently from 0.859 to 0.872, and hallucination declines
from 0.123 to 0.113.  The model is in a consensus-building phase: each
additional step reinforces the reasoning direction without introducing
significant noise.

At step~5, trajectory value dips to 0.897 and hallucination rises to 0.156, 
the first clear inflection point.  
Beyond step~6, the decline accelerates: 
trajectory value drops to 0.649, 
hallucination surges to 0.392, and answer score falls to 0.656. 
Long trajectories do not continue to improve, they may accumulate errors.

\subsection{Trajectory Generation Statistics}
\label{sec:gen_stats}

These numbers point to four steps as a practical upper bound for medical reasoning trajectories.  
Clinical diagnoses typically require a small number of key inferences 
(symptom interpretation, differential narrowing, evidence confirmation, conclusion) 
and extending beyond this natural length introduces noise faster than signal. 
For trajectory construction, this suggests a simple rule:
generate 3--4 steps, evaluate quality at step~4, and discard or
truncate chains that extend further without clear justification.


\begin{table}[htbp]
  \centering
  \caption{Trajectory clusters from the generation pipeline.}
  \label{tab:clusters}
  \small
  \scalebox{0.9}{
  \begin{tabular}{@{}lrrrrr@{}}
    \toprule
    Cluster & Proportion & Traj. Value & Match Score & Hal. \\
    \midrule
    High-quality effective & 82.0\% & 0.999 & 0.979 & 0.029 \\
    Coherent but wrong     & 12.4\% & 0.942 & 0.371 & 0.425 \\
    Low-quality invalid    &  5.5\% & 0.131 & 0.175 & 0.826 \\
    \bottomrule
  \end{tabular}
  }
\end{table}

The full generation pipeline produced 87k valid trajectories comprising
349k step instances across seven injection modes.
KMeans clustering with elbow-method selection of k on step-level features 
(step score, token overlap with clinical points, neighbor coherence, position ratio, 
trajectory value, match score, logical coherence, and evidence score) 
partitions these into 3 groups (Table~\ref{tab:clusters}):

The majority (82\%) are high-quality effective trajectories.
The ``coherent but wrong'' cluster (12.4\%) is the most concerning: 
these trajectories score well on coherence and structure yet arrive at the wrong answer, 
with a hallucination ratio (0.425) far above the high-quality group (0.029). 
They represent the failure mode that is hardest to detect: 
the reasoning looks right but rests on fabricated medical knowledge.

The low-quality invalid cluster (5.5\%) is small but clearly
distinguishable: trajectory value collapses to 0.131 and hallucination
reaches 0.826.  These are the trajectories that the step-level filter
(Section~\ref{sec:filtering}) is designed to catch and exclude from
training data.


\subsection{Validation by Clinical Expert Evaluation}
\label{sec:expert_eval}

\begin{figure}[t]
  \centering
  \subfloat[]{\includegraphics[width=\columnwidth]{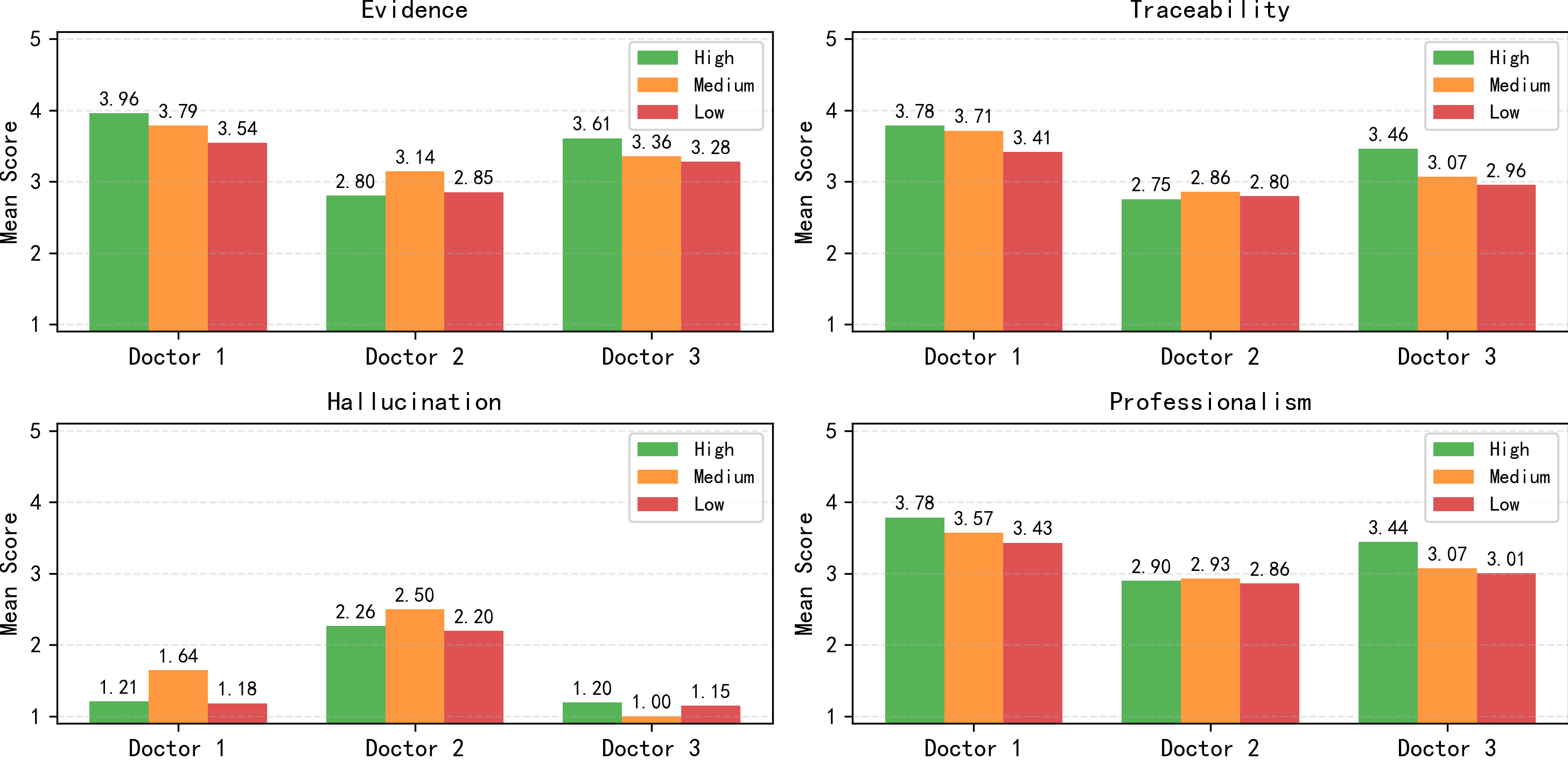}\label{fig:quality_comparison}}\\
  \vspace{-0.5em}
  \subfloat[]{\includegraphics[width=\columnwidth]{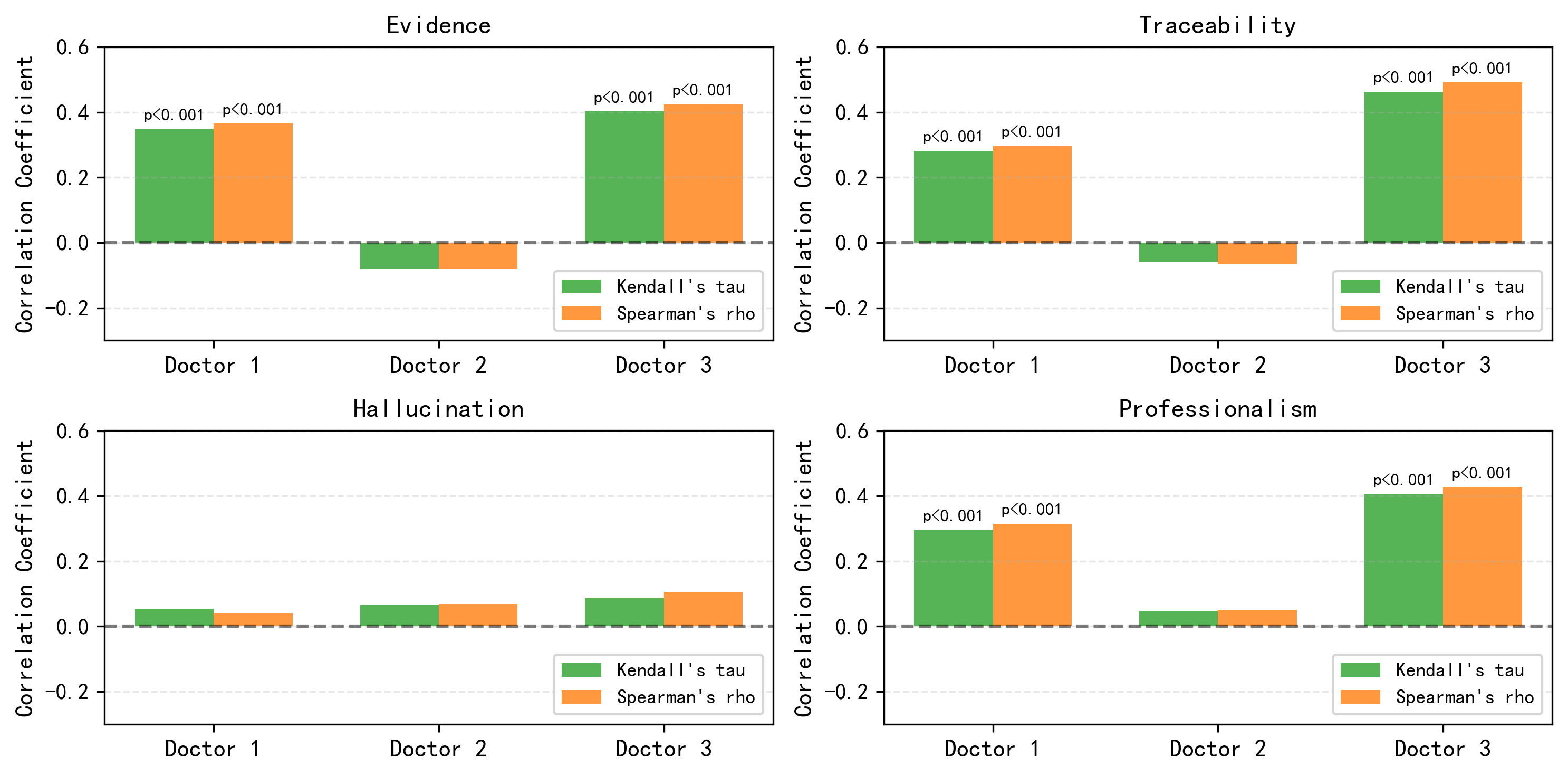}\label{fig:rank_correlation}}
  \vspace{-0.5em}
  \caption{Expert evaluation validation. (a) Mean expert scores stratified by trajectory quality level: trajectories with higher trajectory value received higher scores across all three experts, confirming that the automated quality metric aligns with clinical judgment. (b) Rank correlation between trajectory value and expert scores within each question.}
\end{figure}

Validated on CECMed, Figure~\ref{fig:quality_comparison} presents the mean expert scores stratified by trajectory quality level.
Across all 3 Doctors, trajectories with higher trajectory value consistently received higher scores on evidence support, reasoning traceability, and overall professionalism, while hallucination scores decreased with increasing quality level.
This monotonic relationship confirms that the proposed trajectory value metric captures clinically meaningful quality distinctions.
Notably, the separation between high- and low-quality trajectories was most pronounced for Doctor~3, who exhibited the highest inter-rater agreement with the automated scores.

Figure~\ref{fig:rank_correlation} further quantifies the rank-order agreement between trajectory value and expert scores within each question.
For each doctor-dimension pair, we computed Kendall's $\tau$ and Spearman's $\rho$ measuring whether the ordering of trajectories by trajectory value matches the ordering by expert scores.
For Doctor~1 and Doctor~3, the rank correlations were consistently positive and statistically significant ($p<0.001$) on evidence support, reasoning traceability, and overall professionalism, with Kendall's $\tau$ ranging from 0.28 to 0.46 and Spearman's $\rho$ from 0.30 to 0.49.
The perfect rank concordance rate (i.e., the trajectory value ordering exactly matched the expert score ordering) ranged from 57\% to 83\% across dimensions for these two doctors.
Doctor~2 showed weaker correlations, suggesting a more idiosyncratic evaluation style that diverged from both the automated metric and the other two experts.
The hallucination dimension exhibited lower correlations across the experts, this could be because detecting hallucinations is a more challenging task than evaluating evidence support, reasoning traceability, or professionalism, as it requires identifying specific factual errors rather than assessing the overall quality or reading experience, and judging whether it includes supporting information.

\subsection{Case Study: Expert Evaluation in Practice}
\label{sec:case_study}

To illustrate how the automated trajectory quality metric aligns with clinical expert judgment, we present a representative case from the expert evaluation study.
For a 95-year-old male patient with chest pain and multiple comorbidities, three trajectories were generated at different quality levels: high ($V_{\text{traj}} = 0.9978$), medium ($V_{\text{traj}} = 0.4813$), and low ($V_{\text{traj}} = 0.1419$).
The high-quality trajectory received the highest mean expert score (3.67 out of 5) with zero hallucination, while the medium and low quality trajectories received lower scores (2.67 and 3.33, respectively) and were flagged for logical or factual errors by multiple experts.
This case demonstrates that the proposed trajectory value metric effectively distinguishes clinically meaningful quality differences, though inter-rater disagreement persists, highlighting the subjective nature of clinical quality assessment.
The detailed case study visualization is provided in Appendix~\ref{app:casestudy}.

\section{Limitations}
\label{sec:limitations}

In this work, the trajectory generation and evaluation pipeline generally relies on LLM-based agent.
Both steps may inherit the biases and failure modes of their underlying models.
Hallucinated medical knowledge can enter the trajectory pool at generation time and
pass through the evaluator undetected when the hallucination is internally coherent.
Although our human expert evaluation study validates the automated metric, it is limited to 300 CECMed samples evaluated by 3 experts,
and the hallucination dimension exhibits lower inter-rater agreement than other dimensions,
suggesting that more fine-grained assessment methods are needed to improve the reliability of hallucination detection.

The quality-aware training in this work is limited to context learning and supervised fine-tuning.
Using the trajectory value as a reward signal for reinforcement fine-tuning could be the next step for the further study.
This work demonstrates that trajectory quality signals can effectively guide model behavior through in-context examples alone, 
achieving significant improvements without weight updates.
The trajectory value assessment method and step-level analysis provide the necessary infrastructure for other artificial intelligence approaches,

The error injection modes are designed to isolate specific fault types.
Real clinical reasoning errors could be more complex and nuanced.
The controlled setting establishes causal links between fault type and quality drop,
but the mapping from injection mode to natural error is approximate.
This work validates that structural disruptions to reasoning chains cause more severe quality degradation than local factual errors, 
future work can extend the injection catalog with domain-specific error patterns derived from clinical case reviews.

\section{Conclusion}
\label{sec:conclusion}

This work proposes MedTraj,
a framework that treats reasoning trajectories as primary objects for modeling the progressive refinement of clinical evidence toward diagnostic decisions.
By capturing the state transitions among medical concepts,
MedTraj makes the implicit reasoning chain of a medical agent explicit and inspectable.
The framework introduces a multi-dimensional quality evaluation scheme with controlled error injection,
a composite trajectory value metric,
and a step-level filtering mechanism that identifies key driver and detrimental steps.
Experiments across three open-ended medical QA datasets demonstrate that
trajectory context consistently improves reasoning quality,
with logical coherence gains of +0.029 to +0.041 across all tasks.
Quality-weighted context achieves 73.8\% correctness on CECMed with an 87\% reduction in hallucination ratio,
while step-level analysis reveals that 15.3\% of steps act as key drivers
and the number of reasoning steps forming a practical upper bound for medical reasoning length may be as few as four.
This work focuses on establishing the feasibility of trajectory quality feedback as a training signal,
providing practical guidelines for constructing and evaluating reasoning trajectories of LLM-based medical agents.
Validation by three clinical experts confirms that the proposed trajectory value metric generally aligns with human clinical judgment.
Future work could integrate the trajectory value as a reward signal for policy optimization,
expand validation to more clinical domains and model families,
and incorporate human expert evaluation within the pipeline.


\bibliographystyle{IEEEtran}
\bibliography{references}

\newpage

\onecolumn

\appendix

\section{Main Results Visualization}
\label{app:visual}

\begin{figure*}[htbp]
  \centering
  \includegraphics[width=0.88\textwidth]{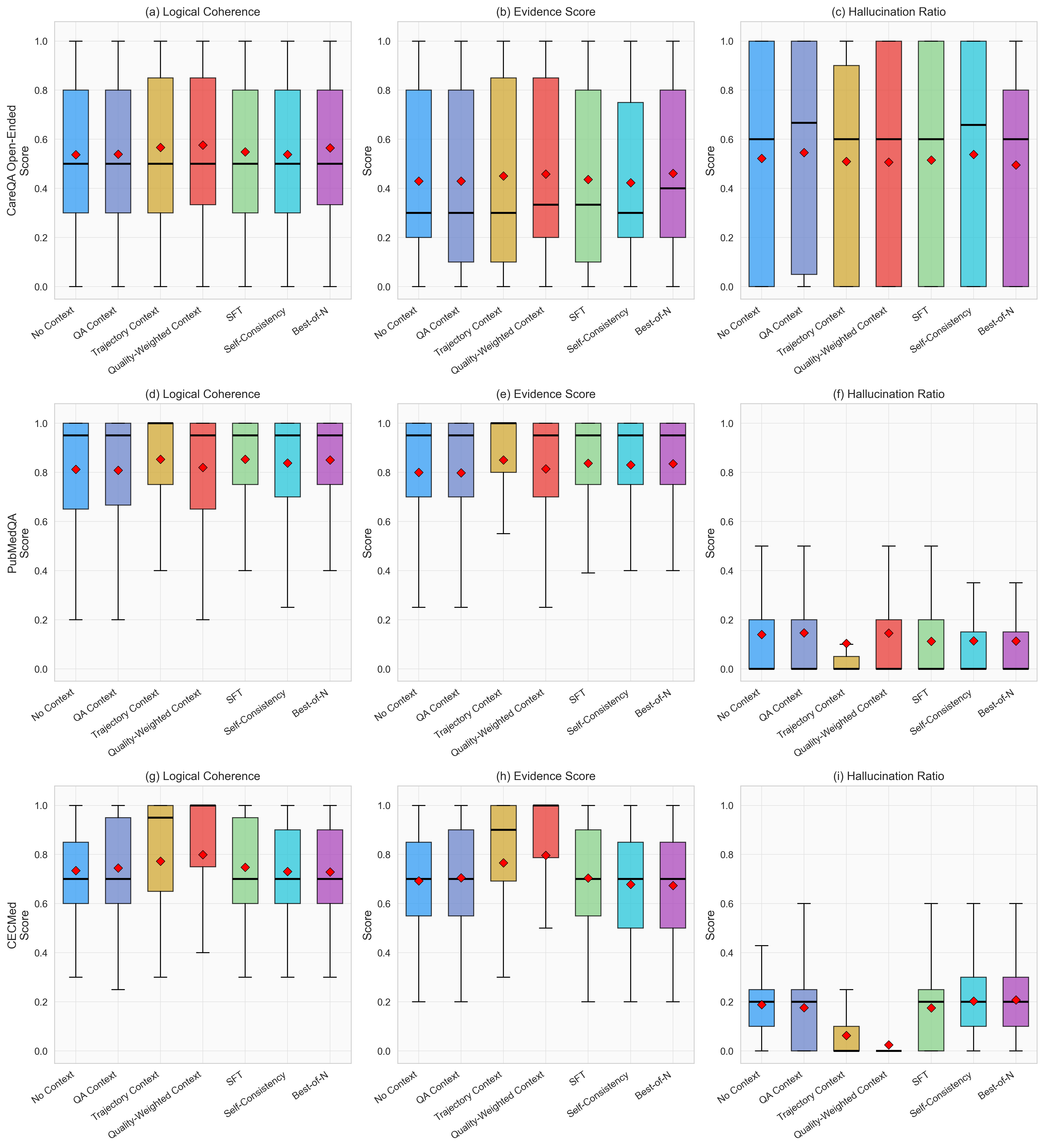}
  \vspace{-0.5em}
  \caption{Distributions of correctness, coherence,
    evidence score, and hallucination across all methods and
    datasets. (red diamond = mean)}
  \label{fig:appendix_boxplots}
\end{figure*}

\begin{figure*}[htbp]
  \centering
  
  \subfloat[]{\includegraphics[width=0.43\textwidth]{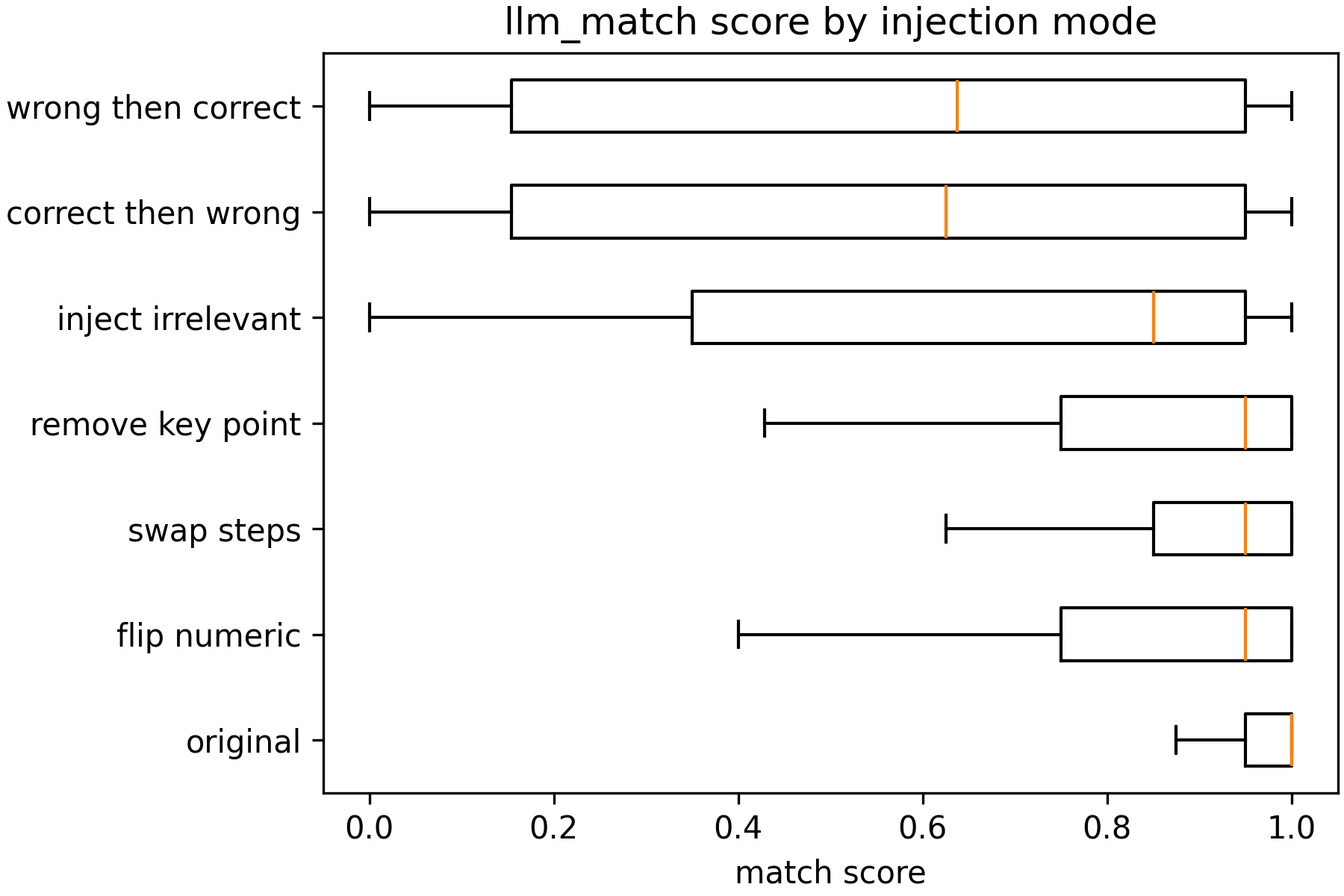}\label{fig:box_match}}
  \hfill
  \subfloat[]{\includegraphics[width=0.43\textwidth]{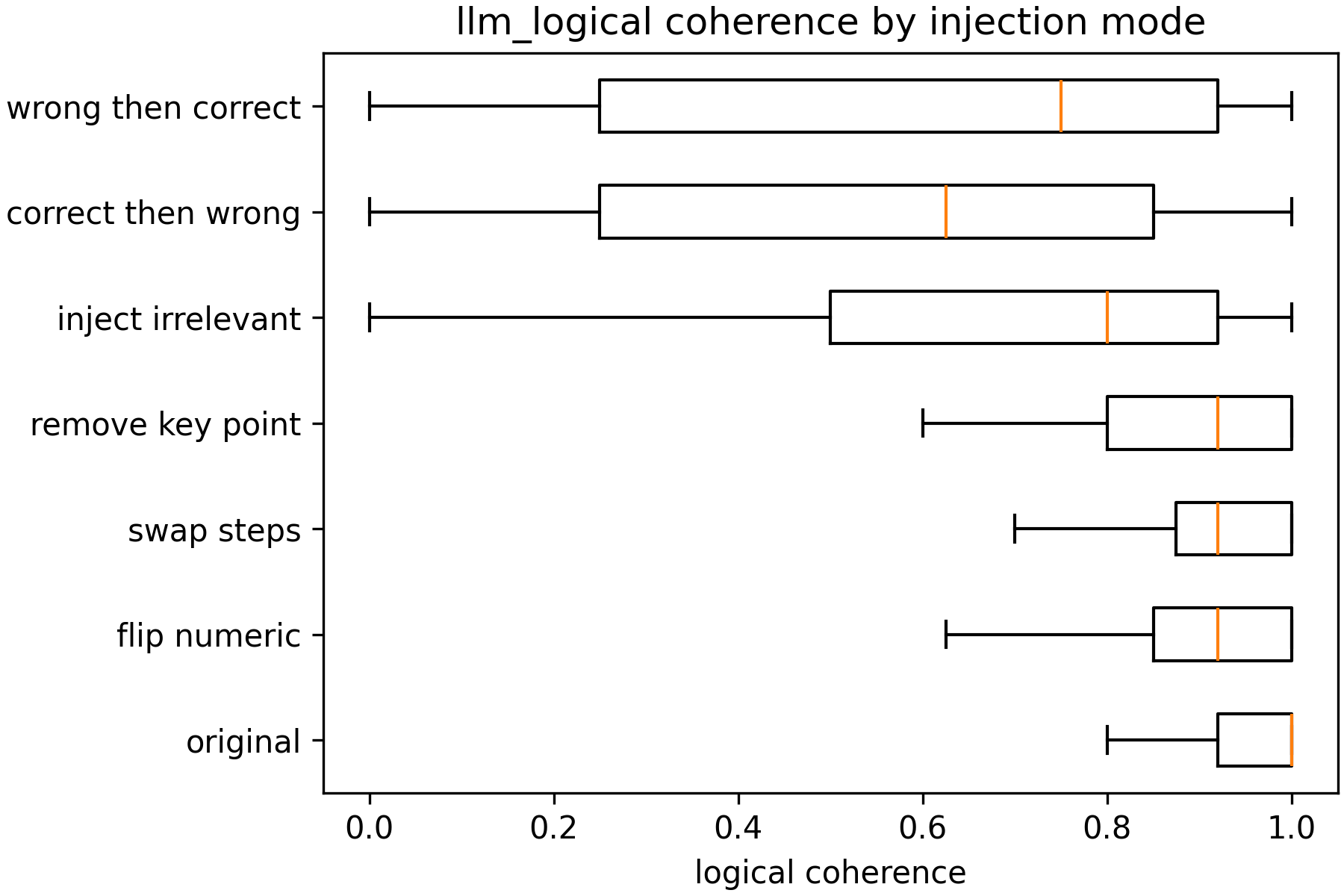}\label{fig:box_coherence}}

  \subfloat[]{\includegraphics[width=0.43\textwidth]{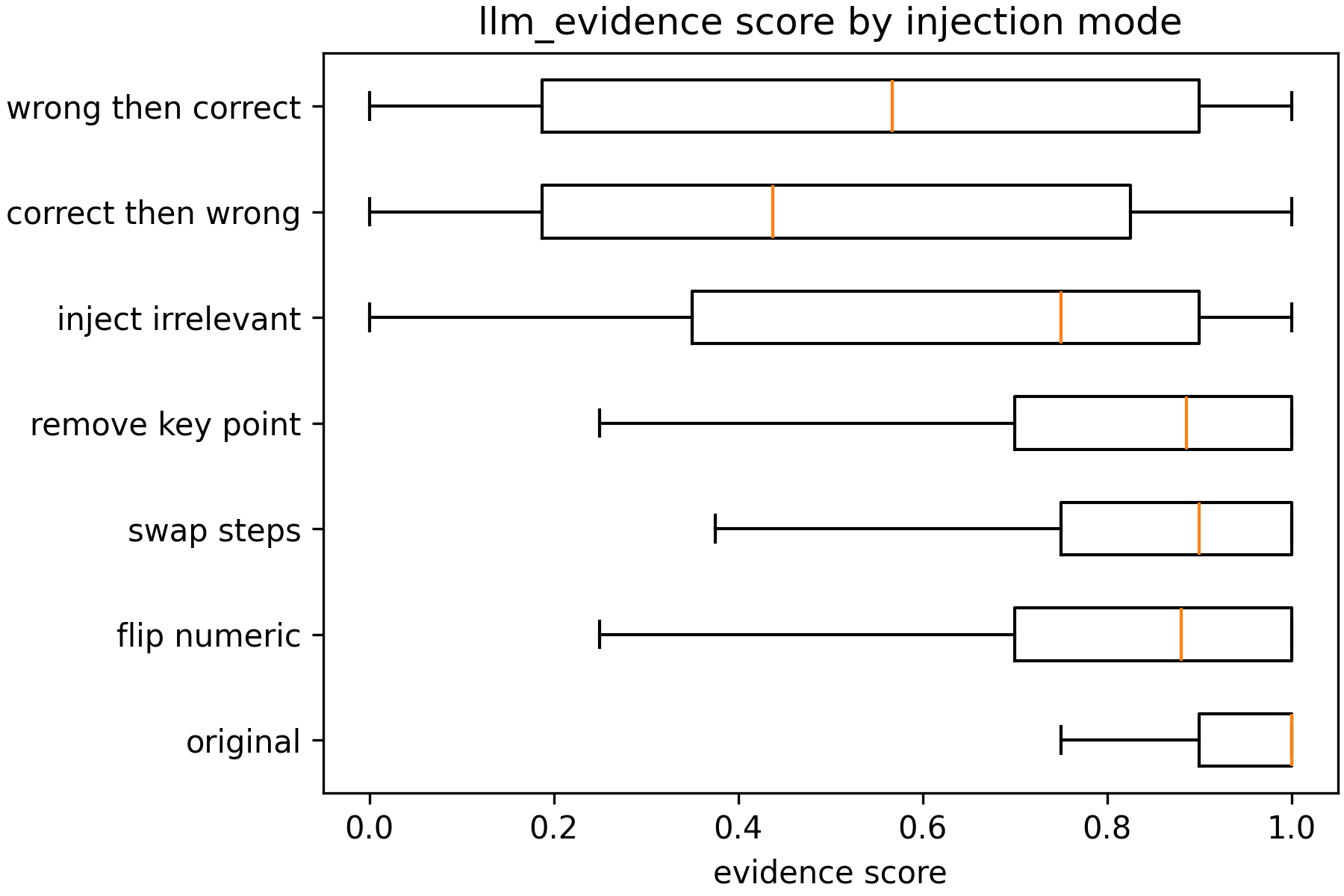}\label{fig:box_evidence}}
  \hfill
  \subfloat[]{\includegraphics[width=0.43\textwidth]{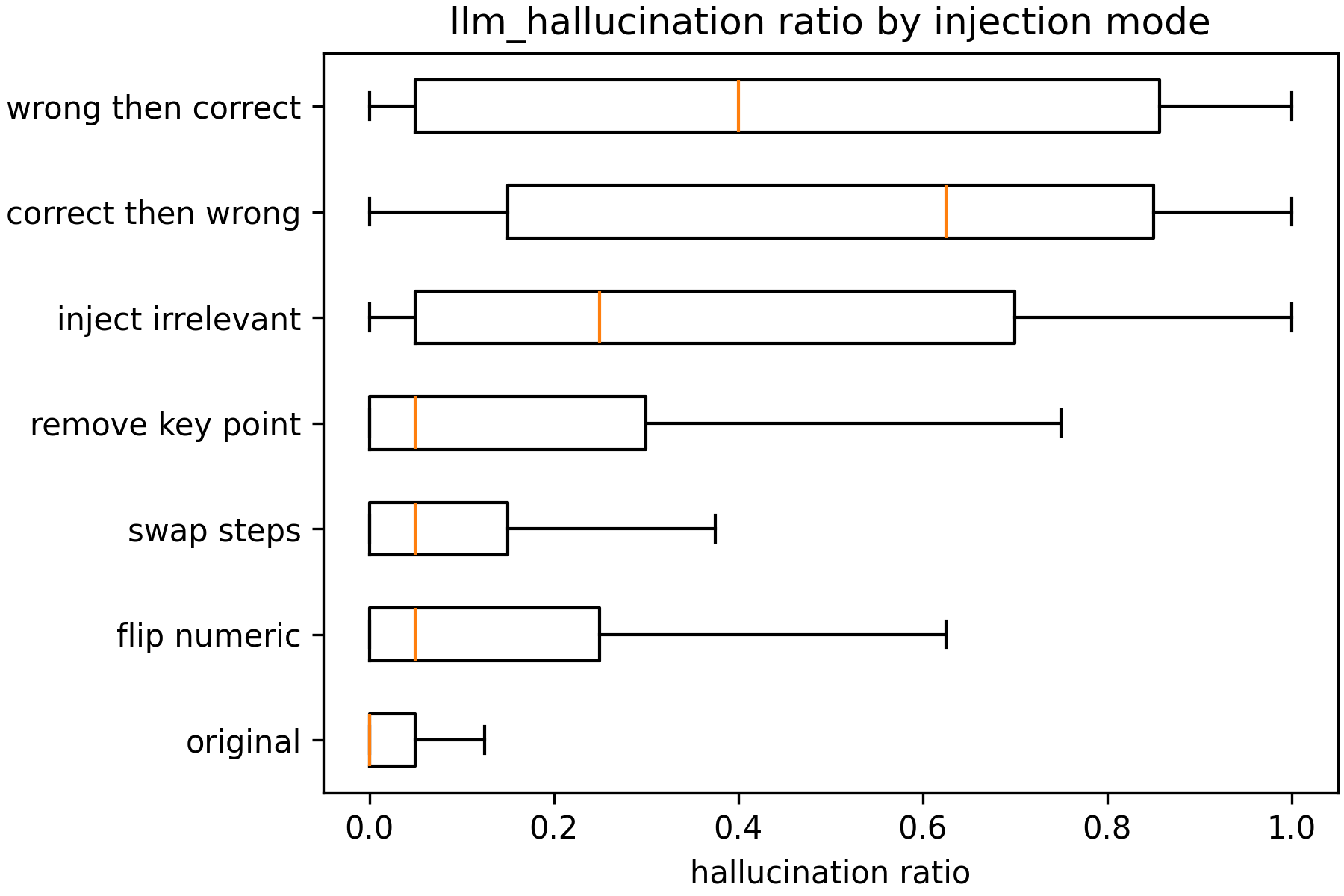}\label{fig:box_hallucination}}

  \subfloat[]{\includegraphics[width=0.43\textwidth]{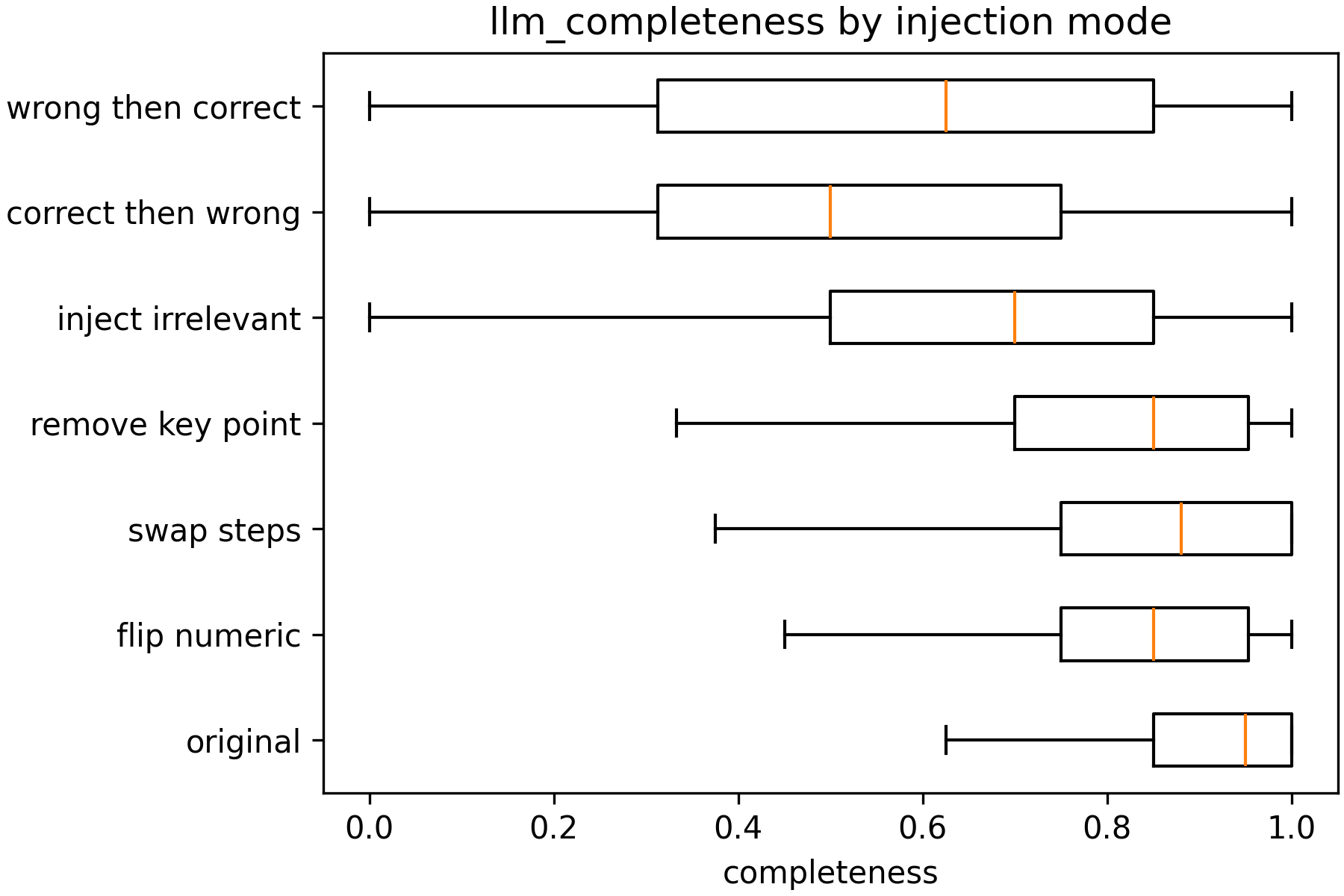}\label{fig:box_completeness}}
  \hfill
  \subfloat[]{\includegraphics[width=0.43\textwidth]{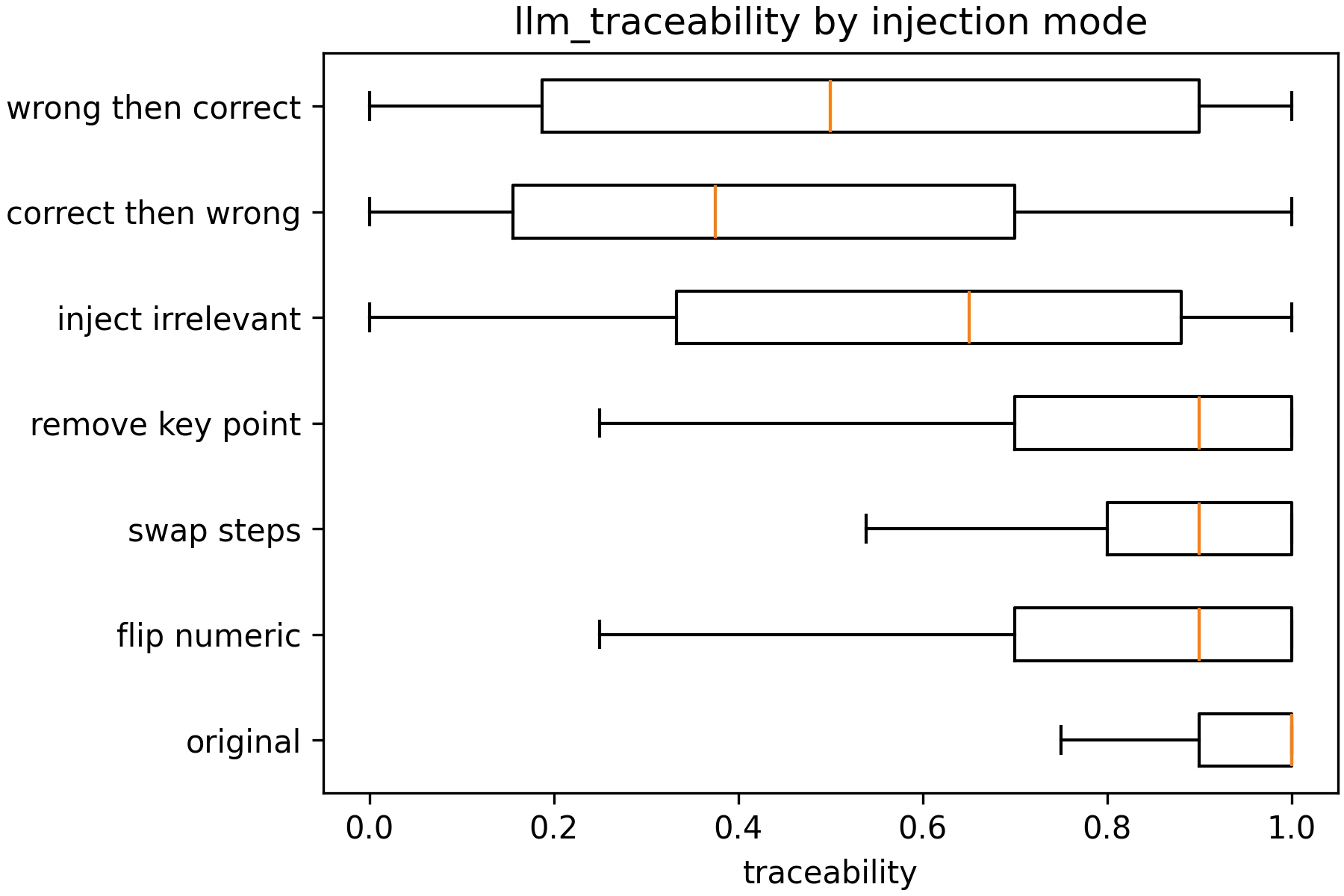}\label{fig:box_traceability}}

  \subfloat[]{\includegraphics[width=0.43\textwidth]{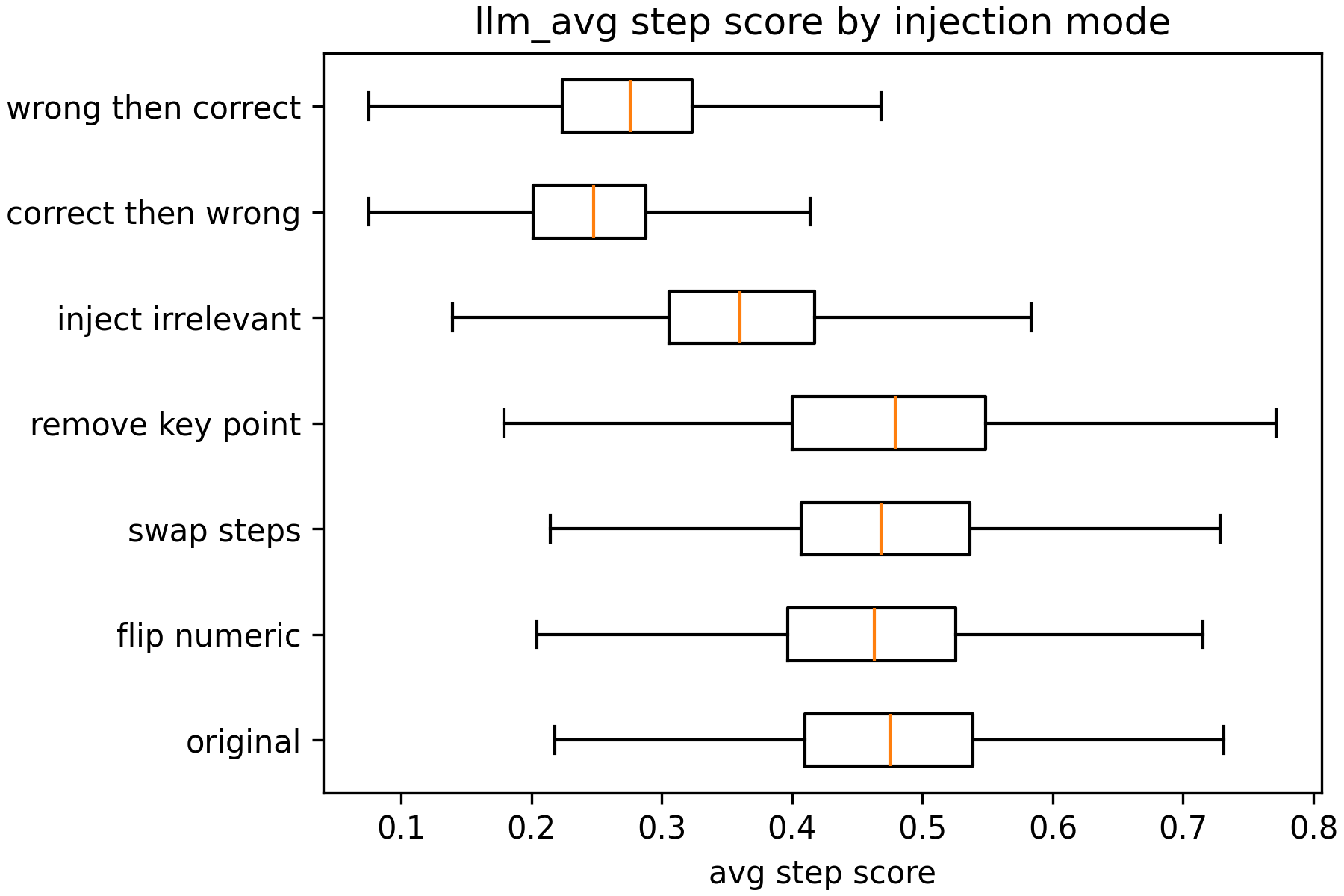}\label{fig:box_avg_step_score}}
  \hfill
  \subfloat[]{\includegraphics[width=0.43\textwidth]{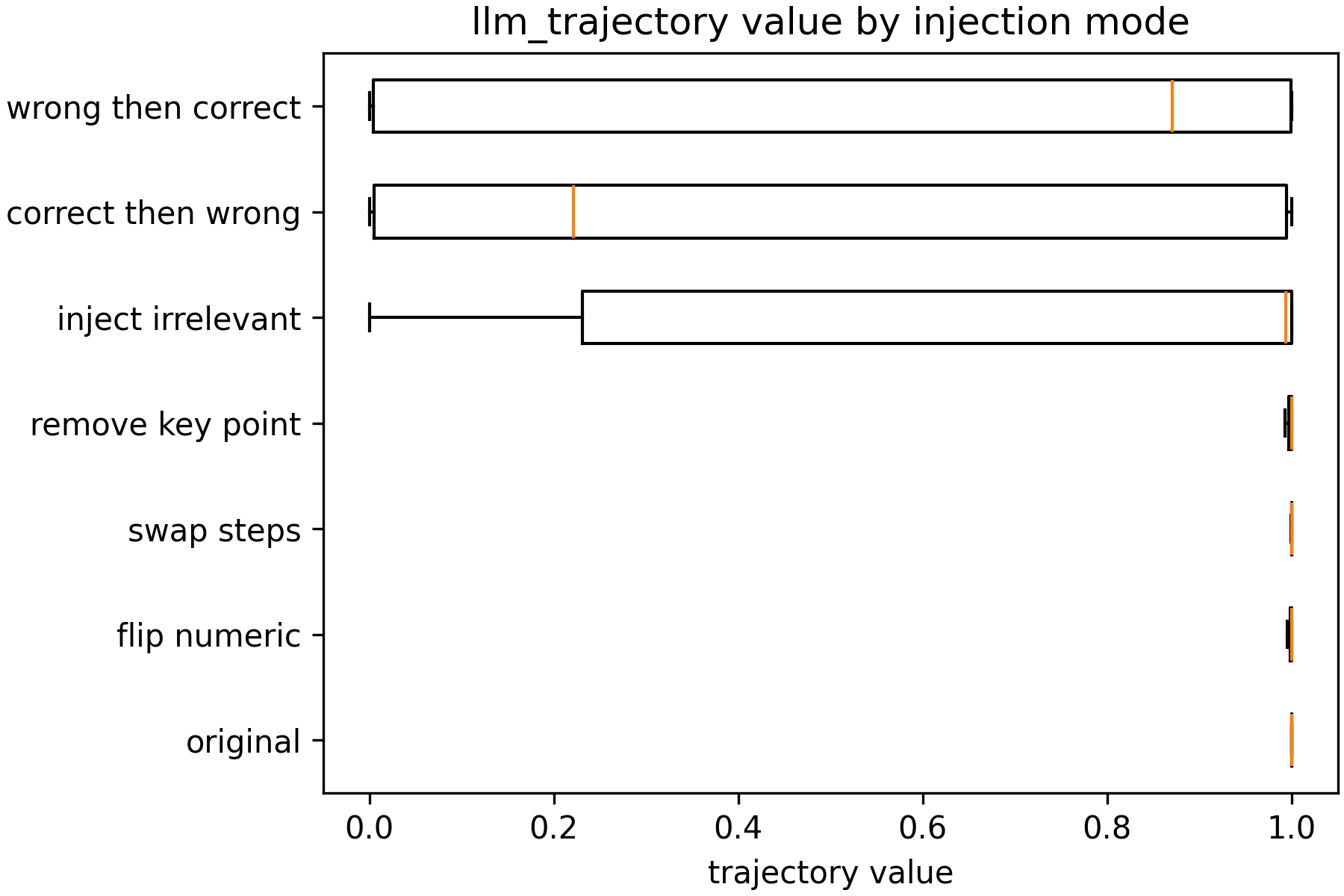}\label{fig:box_trajectory_value}}

  \caption{Distributions of evaluation dimensions under each injection mode.}
  \label{fig:injection_mode_boxplots}
\end{figure*}

Figure~\ref{fig:appendix_boxplots} shows the distributions
of all four evaluation metrics across the seven methods and three
datasets, complementing the mean values reported in
Table~\ref{tab:main_results}. Figure~\ref{fig:injection_mode_boxplots} shows the per-sample
distributions of all seven evaluation dimensions under each injection
mode, complementing the growth curves in Figure~\ref{fig:error_injection_curves}.

\newpage

\definecolor{promptbg}{RGB}{245, 248, 252}
\definecolor{promptborder}{RGB}{137,158,193}
\definecolor{sectionbg}{RGB}{230, 240, 250}
\definecolor{codebg}{RGB}{250, 250, 250}
\definecolor{casebg}{RGB}{248, 252, 245}
\definecolor{caseborder}{RGB}{220, 240, 220}
\definecolor{casehigh}{RGB}{220, 240, 220}
\definecolor{casemed}{RGB}{255, 248, 220}
\definecolor{caselow}{RGB}{240,233,233}

\newtcolorbox{promptbox}[1][]{
  colback=promptbg,
  colframe=promptborder,
  boxrule=0.8pt,
  arc=3pt,
  left=8pt, right=8pt, top=6pt, bottom=6pt,
  breakable,
  title=#1,
  fonttitle=\bfseries\small,
  fontupper=\small,
  fontlower=\small,
  coltitle=black,
}

\newtcolorbox{casebox}[1][]{
  colback=casebg,
  colframe=caseborder,
  boxrule=0.8pt,
  arc=3pt,
  left=8pt, right=8pt, top=6pt, bottom=6pt,
  breakable,
  title=#1,
  fonttitle=\bfseries\small,
  fontupper=\small,
  fontlower=\small,
  coltitle=black,
}

\section{Case Study}
\label{app:casestudy}

\begin{figure*}[htbp]
  \centering
  \includegraphics[width=\textwidth]{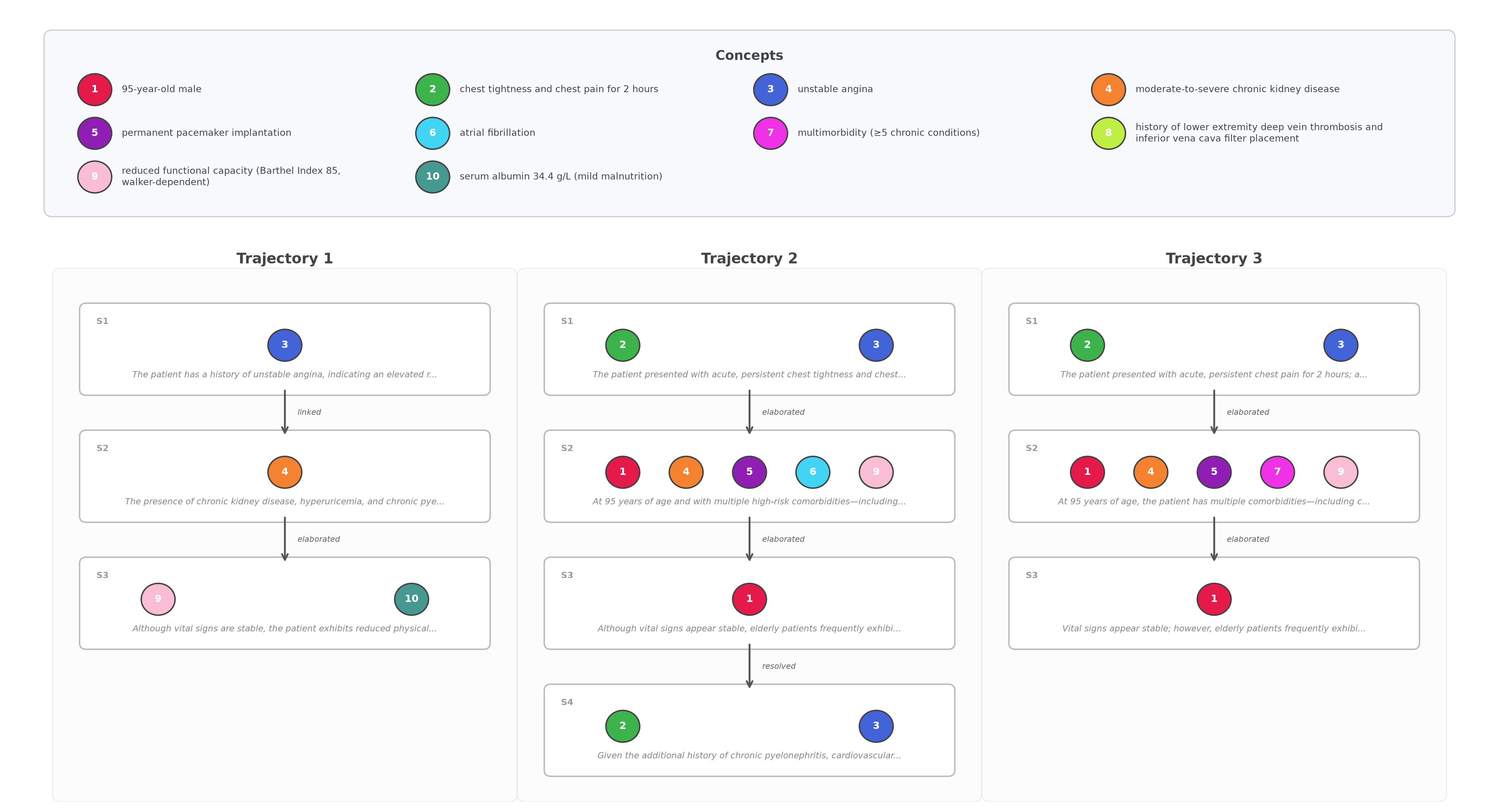}
  \vspace{-0.5em}
  \caption{Case Study: Trajectories Visualization.}
  \label{fig:traj-case-visualize}
\end{figure*}

\begin{casebox}[Case Study: Expert Evaluation of Three Quality Levels]
\textbf{Question:} 
The following is the admission clinical information of an elderly patient. Please comprehensively assess the severity of this patient's condition based on these details.
A 95-year-old male was admitted to the emergency department with "chest tightness accompanied by chest pain for 2 hours." The patient has a history of long-term heavy alcohol consumption (one or more times daily) and has quit smoking for over one year. Height: 175 cm, weight: 78 kg, BMI: 25.4 (overweight). Past medical history includes moderate-to-severe chronic kidney disease, permanent pacemaker implantation (implanted 13 years ago, battery replaced 4 years ago), hyperlipidemia, hypertension, hyperuricemia, peptic ulcer, atrial fibrillation with right-sided facial paralysis, benign prostatic hyperplasia, and chronic bronchitis. Comorbidities include sick sinus syndrome, unstable angina, chronic pyelonephritis, and grade 1 hypertension. The patient is on long-term antihypertensive medications, lipid-lowering agents, and proton pump inhibitors, with a positive allergy history.
Two hours prior to admission, the patient developed sudden chest tightness and chest pain that persisted without relief, prompting emergency evaluation. The patient has a history of deep vein thrombosis in the lower extremities and underwent inferior vena cava filter placement. Surgical history includes appendectomy (40 years ago), bilateral first metatarsophalangeal joint correction (18 years ago), left second toe proximal interphalangeal joint fusion (17 years ago), and cataract surgery (9 years ago). Currently, the patient ambulates with a walker, reports easy fatigability, and has reduced activity level. Barthel Index: 85, Morse Fall Risk Assessment: 45.
Vital signs are stable: temperature 36.5°C, blood pressure 137/66 mmHg, heart rate 61 bpm, respiratory rate 18 breaths/min, pulse oximetry 96\%. Laboratory findings: hemoglobin 133 g/L, albumin 34.4 g/L (mildly decreased), urine white blood cells 5.06/$\mu$L, red blood cells 4.11/$\mu$L. Cardiac enzyme panel shows no significant abnormalities. Blood glucose reading is 0 (requires verification). Coagulation function is essentially normal. Comprehensive geriatric assessment indicates decreased mobility and fatigue symptoms, with no pain or fall risk.
Please determine the severity level of this patient's condition: Mild, Moderate, or Severe? 

Reference answer: \textbf{Medium}.

\vspace{0.5em}
\noindent\begin{minipage}{\textwidth}
\colorbox{casehigh}{\textbf{Traj 1}} \; $V_{\text{traj}} = 0.9978$, Answer: Medium\\
\textit{Reasoning:} 
  The patient has a history of unstable angina, indicating risk of acute coronary syndrome, which represents a serious cardiovascular condition.
  The patient has chronic kidney disease, hyperuricemia, and chronic pyelonephritis, suggesting multi-organ system involvement and complex clinical presentation.
  Although vital signs are stable, decreased mobility, significant fatigue, and fall risk indicate poor overall functional status.
  Answer: Moderate\\
\textit{Automated Scores:} Coherence=0.50, Evidence=0.75, Hallucination=0.00, Completeness=0.60, Traceability=0.50\\
\textit{Expert Overall (D1/D2/D3):} \textbf{5/4/2} (Mean=3.67)
\end{minipage}

\vspace{0.8em}
\noindent\begin{minipage}{\textwidth}
\colorbox{casemed}{\textbf{Traj 2}} \; $V_{\text{traj}} = 0.4813$, Answer: Severe\\
\textit{Reasoning:} The patient presented with sudden onset of persistent chest tightness and chest pain for 2 hours. Although cardiac enzymes were not significantly elevated, the history of unstable angina warrants high suspicion for acute coronary syndrome, representing a moderate-to-high risk cardiovascular event.
  At 95 years of age with multiple high-risk factors (chronic kidney disease, atrial fibrillation, hypertension, hyperlipidemia, history of thrombosis), comprehensive geriatric assessment shows decreased mobility and fatigue, indicating poor physiological reserve and significantly increased risk of complications.
  Vital signs appear stable on the surface, but elderly patients often present with occult shock, requiring close hemodynamic monitoring. The condition is complex and dangerous.
  Normal coagulation function contradicts the history of inferior vena cava filter placement, necessitating verification of coagulation test accuracy. Blood glucose reading of 0 (requires verification) may affect interpretation of cardiac injury markers, suggesting potential diagnostic uncertainty.
  Abnormal urinalysis (white blood cells 5.06/$\mu$L) suggests possible urinary tract infection, but given the history of chronic pyelonephritis, cardiovascular emergency should be prioritized over infection. Overall assessment: condition severity is "Severe."
  Answer: Severe\\
\textit{Automated Scores:} Coherence=0.44, Evidence=0.56, Hallucination=0.63, Completeness=0.38, Traceability=0.25\\
\textit{Expert Overall (D1/D2/D3):} \textbf{3/3/2} (Mean=2.67)
\end{minipage}

\vspace{0.8em}
\noindent\begin{minipage}{\textwidth}
\colorbox{caselow}{\textbf{Traj 3}} \; $V_{\text{traj}} = 0.1419$, Answer: Severe\\
\textit{Reasoning:} The patient presented with sudden onset of persistent chest pain for 2 hours. Although cardiac enzymes were not significantly elevated, the history of unstable angina suggests moderate-to-high risk probability of acute coronary syndrome.
  The patient is 95 years old with comorbidities including chronic kidney disease, atrial fibrillation, hypertension, hyperlipidemia, history of thrombosis, and decreased mobility, indicating poor physiological reserve and significantly increased risk of complications.
  Vital signs appear stable on the surface, but elderly patients often present with occult shock, requiring close hemodynamic monitoring. The condition assessment should be considered high-risk.
  Abnormal urinalysis may suggest urinary tract infection, but given the history of chronic pyelonephritis, cardiovascular emergency should be the primary concern. Overall assessment: condition severity is "Severe."
  Answer: Severe\\
\textit{Automated Scores:} Coherence=0.40, Evidence=0.30, Hallucination=0.70, Completeness=0.20, Traceability=0.20\\
\textit{Expert Overall (D1/D2/D3):} \textbf{4/3/3} (Mean=3.33)
\end{minipage}
\end{casebox}

\newpage

\section{Prompt Templates}
\subsection{Answer Generation}
\label{app:ans-prompt}

\begin{promptbox}[Persona]
\begin{verbatim}
You are a professional, board-certified clinical physician with 
excellent medical knowledge and strong step-by-step clinical reasoning. 
Answer as an expert clinician: identify salient findings, 
consider the most likely diagnoses or decisions, 
and justify your recommendation logically.
\end{verbatim}
\end{promptbox}

\begin{promptbox}[Instructions]
\begin{verbatim}
Instructions (REQUIRED):
1) Output ONLY a single valid JSON object and nothing else.
2) For multiple-choice questions, include keys: ``choice''
   (single uppercase letter, e.g. ``A'') and ``explanation_steps''
   (an array of short strings).
3) For open-ended questions, include keys: ``answer'' (short final 
    answer) and ``explanation_steps'' (an array of short strings).
4) Each ``explanation_steps'' entry must be a numbered step
   (start with ``1.'' or ``Step 1:'') describing one logical
   point in the clinician's reasoning.
5) Provide between 2 and 5 steps; each step should be concise 
    (1-2 sentences).
6) Do NOT include any text outside the JSON object (no preface, no commentary).
\end{verbatim}
\end{promptbox}

\begin{promptbox}[Open-Ended Answer Generation Prompt]
\begin{verbatim}
Now answer the following clinical question and provide
numbered reasoning steps (as JSON):
{question}
\end{verbatim}
\end{promptbox}

\subsection{Trajectory Parsing}

\begin{promptbox}[Trajectory Parsing Prompt]
\begin{verbatim}
Parse the following clinical reasoning trajectory into JSON
with keys: clinical_points (list), evidence (list),
reasoning_steps (list of {step_id, content, sub_reasoning}),
final_result ({diagnosis, treatment, prognosis}).
Output only JSON. Trajectory:
{free_form_trajectory_text}
\end{verbatim}
\end{promptbox}

\subsection{LLM Trajectory Evaluation}

\begin{promptbox}[Judge Prompt]
\begin{verbatim}
You are a clinical reasoning quality evaluator. Provided below are: 
the question, the reference answer, the model prediction (if any), 
and the parsed trajectory
(clinical_points, reasoning_steps, final_result).

Based only on the parsed trajectory and the reference answer, 
return a single JSON object and nothing else (no extra text). 
The JSON MUST contain the following fields:
- is_correct: boolean or null (true if the prediction
  matches the reference; null if no prediction)
- match_score: float 0.0-1.0 (similarity to reference)
- logical_coherence: float 0.0-1.0 (coherence between
  reasoning steps and clinical points)
- evidence_score: float 0.0-1.0 (degree to which claims
  are supported by clinical points or evidence)
- hallucination_ratio: float 0.0-1.0 (proportion of
  assertions not grounded in clinical points)
- completeness: float 0.0-1.0 (how complete the trajectory
  is: points, steps, and conclusion)
- traceability: float 0.0-1.0 (how well the final
  conclusion is traceable to the reasoning steps)

Question: {question}
Reference: {reference}
Prediction: {pred}

Parsed trajectory:
Clinical points: {cp_text}

Reasoning steps:
{rs_text}

Final result - diagnosis: {fr_diag}
Treatment: {fr_treat}

Return only the JSON object.
or other text.
\end{verbatim}
\end{promptbox}

\subsection{Context Learning Baselines}

\begin{promptbox}[No Context baseline]
\begin{verbatim}
Now answer the following clinical question and provide
numbered reasoning steps (as JSON):
{question}
\end{verbatim}
\end{promptbox}

\begin{promptbox}[QA Context Prompt]
\begin{verbatim}
Here are some example clinical questions and answers
for reference:

--- Example 1 ---
Question: {ex1_question}
Answer: {ex1_answer}

--- Example 2 ---
Question: {ex2_question}
Answer: {ex2_answer}
... (k QA pair examples)

--- Now answer the following question ---
{question}
\end{verbatim}
\end{promptbox}

\begin{promptbox}[Trajectory Context Prompt]
\begin{verbatim}
Here are some example clinical reasoning trajectories
for reference. Pay attention to the step-by-step
reasoning process:

--- Example 1: Clinical Reasoning ---
Question: {ex1_question}
Reasoning steps:
  1. {ex1_step1}
  2. {ex1_step2}
  ...
Answer: {ex1_answer}
... (k high-quality trajectories)

Now answer the following question with similar reasoning
{question}
\end{verbatim}
\end{promptbox}

\begin{promptbox}[Quality-Weighted Context Prompt]
\begin{verbatim}
Here are example clinical reasoning trajectories with
quality scores. Higher quality examples demonstrate good
reasoning patterns; lower quality examples show common
pitfalls to avoid. Use the quality scores as a guide
for what constitutes good reasoning:

--- Example 1: Clinical Reasoning
    (Quality: High, Value=0.85, Match=0.90) ---
Question: {ex1_question}
Reasoning steps:
  1. {ex1_step1}
  ...
Answer: {ex1_answer}
Note: This example demonstrates high-quality clinical
reasoning. Follow this pattern.

--- Example 2: Clinical Reasoning
    (Quality: Low, Value=0.25, Match=0.30) ---
Question: {ex2_question}
Reasoning steps:
  1. {ex2_step1}
  ...
Answer: {ex2_answer}
Note: This example has lower quality — the reasoning
may contain errors or gaps. Use it as a cautionary
reference.

... (balanced high/low quality, k total)

Now answer the following question, aiming for high-quality reasoning
{question}
\end{verbatim}
\end{promptbox}

\end{document}